\documentclass[10pt,twocolumn,letterpaper]{article}

\usepackage[pagenumbers]{iccv}

\usepackage{color}
\usepackage{xcolor}
\definecolor{deepGreen}{RGB}{0,153,0}
\definecolor{orange}{RGB}{255,125,0}
\def\red#1{\textcolor[rgb]{1,0,0}{#1}}
\def\blue#1{\textcolor[rgb]{0,0,1}{#1}}

\definecolor{sainone}{RGB}{236, 242, 249}
\definecolor{saintwo}{RGB}{255, 230, 204}

\newcommand{\keypoint}[1]{\vspace{0.1cm}\noindent\textbf{#1}\;}
\newcommand{\cut}[1]{}
\usepackage{multirow}
\usepackage{amsmath}
\usepackage{amssymb}
\usepackage{mathtools}
\usepackage{arydshln}
\usepackage{soul}
\usepackage{nicefrac}
\usepackage{booktabs}
\usepackage{stmaryrd}
\usepackage{caption}
\usepackage{graphicx}
\usepackage{colortbl}
\definecolor{gray}{gray}{0.9}
\definecolor{pink}{RGB}{255, 234, 232}
\definecolor{ourPink}{RGB}{248, 206, 204}
\definecolor{existing}{RGB}{176, 227, 230}
\sethlcolor{pink}
\usepackage{wrapfig}
\usepackage{anyfontsize}
\usepackage{graphicx}
\usepackage{booktabs}
\usepackage[accsupp]{axessibility}
\makeatletter
\newcommand\notsotiny{\@setfontsize\notsotiny\@vipt\@viipt}
\makeatother
\makeatletter
\apptocmd\@maketitle{{\myfigure{}\par}}{}{}
\makeatother
\usepackage{capt-of,etoolbox}

\definecolor{cat_level}{HTML}{2D7399}
\definecolor{fine_grained}{HTML}{43AA8B}
\definecolor{part_level}{HTML}{981B1B}
\definecolor{composed}{HTML}{F0A30A}

\definecolor{iccvblue}{rgb}{0.21,0.49,0.74}
\usepackage[pagebackref,breaklinks,colorlinks,allcolors=iccvblue]{hyperref}

\title{\vspace{-10mm}SketchYourSeg: Mask-Free Subjective Image Segmentation via Freehand Sketches\vspace{-8mm}}

\author{Subhadeep Koley\textsuperscript{1,2} \hspace{.2cm} Viswanatha Reddy Gajjala\thanks{Interned with SketchX} \hspace{.2cm} Aneeshan Sain\textsuperscript{1} \hspace{.2cm}  Pinaki Nath Chowdhury\textsuperscript{1}\\ Tao Xiang\textsuperscript{1,2}\hspace{.3cm} Ayan Kumar Bhunia\textsuperscript{1}\hspace{.3cm}  Yi-Zhe Song\textsuperscript{1,2} \\
\textsuperscript{1}SketchX, CVSSP, University of Surrey, United Kingdom.  \\
\textsuperscript{2}iFlyTek-Surrey Joint Research Centre on Artificial Intelligence.\\
{\tt\small \{s.koley, a.sain, p.chowdhury, t.xiang, a.bhunia, y.song\}@surrey.ac.uk}
}

\vspace{-0.7cm}

\newcommand\myfigure{
\centering
\vspace{-1cm}
\captionsetup{type=figure}
    \includegraphics[width=\textwidth]{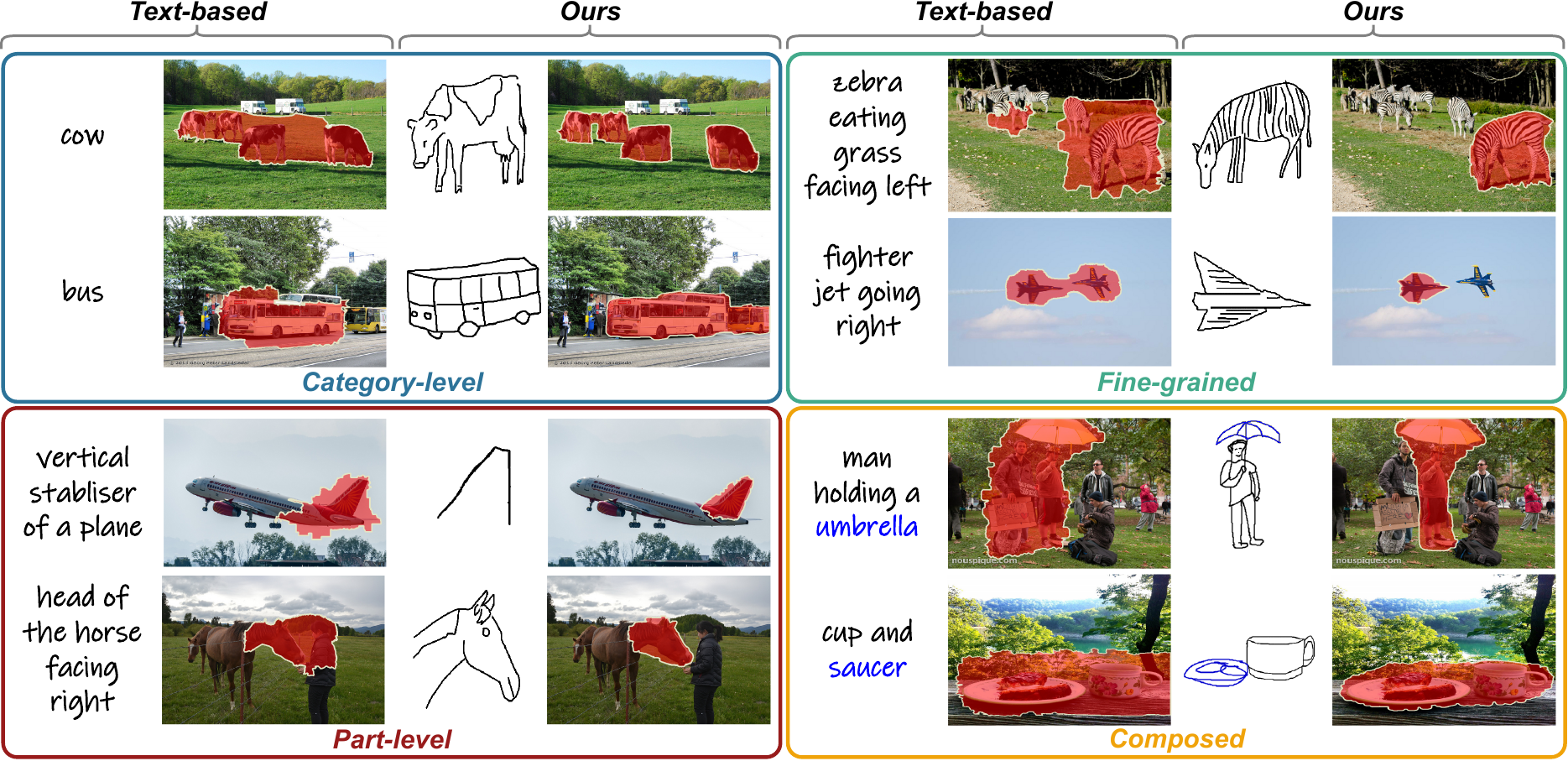}
    \vspace{-0.7cm}
\captionof{figure}{Comparison of a few segmented images by our method with a text-based segmenter (\textit{cf}.\ \cref{sec:competitors}). Utilisation of sketch enables segmentation at multiple granularity levels (\eg, category-level, fine-grained, part-level). While writing granularity-specific textual cues is cumbersome and difficult, with simple freehand sketches, the proposed method segments either ``all'' cows (\ie, \textcolor{cat_level}{category-level}), ``that'' zebra (\ie, \textcolor{fine_grained}{fine-grained}), ``head'' of the horse (\ie, \textcolor{part_level}{part-level}), or even \textcolor{composed}{composed} concepts like a ``cup'' \textit{and} a ``saucer''.}
\label{fig:teaser}
\vspace{+0.25cm}
}

\begin{document}
\maketitle

\vspace{-0.5cm}
\begin{abstract}
We introduce SketchYourSeg, a novel framework that establishes freehand sketches as a powerful query modality for subjective image segmentation across entire galleries through a single exemplar sketch. Unlike text prompts that struggle with spatial specificity or interactive methods confined to single-image operations, sketches naturally combine semantic intent with structural precision. This unique dual encoding enables precise visual disambiguation for segmentation tasks where text descriptions would be cumbersome or ambiguous -- such as distinguishing between visually similar instances, specifying exact part boundaries, or indicating spatial relationships in composed concepts. Our approach addresses three fundamental challenges: (i) eliminating the need for pixel-perfect annotation masks during training with a mask-free framework; (ii) creating a synergistic relationship between sketch-based image retrieval (SBIR) models and foundation models (CLIP/DINOv2) where the former provides training signals while the latter generates masks; and (iii) enabling multi-granular segmentation capabilities through purpose-made sketch augmentation strategies. Our extensive evaluations demonstrate superior performance over existing approaches across diverse benchmarks, establishing a new paradigm for user-guided image segmentation that balances precision with efficiency.

\end{abstract}

\vspace{-0.5cm}
\section{Introduction}
\label{sec:intro}
This paper introduces a novel approach to image segmentation that leverages freehand sketches for \textit{subjective} segmentation across \textit{entire galleries} with a \textit{single} exemplar sketch. Our method, \textit{SketchYourSeg}, enables users to express personalised segmentation intent through simple sketches rather than complex text descriptions or repetitive interactive inputs, whether targeting whole objects, specific parts, or multiple related elements.

Image segmentation has evolved substantially \cite{kirillov2023segment}, with current systems increasingly leveraging foundation models in open-vocabulary settings \cite{mukhoti2023open}. However, a critical question remains: how can users effectively communicate their subjective segmentation intent while efficiently operating across multiple images? While text prompts \cite{xu2022simple} and interactive methods \cite{chen2021conditional, hao2021edgeflow} have gained popularity, they present fundamental limitations for precisely specifying segmentation targets, especially across entire image sets.

To address this challenge, we investigate freehand sketches as a query modality for subjective segmentation. A key question arises: why use sketches when alternatives like text prompts or interactive scribbles exist? Research evidence demonstrates that sketches occupy a unique middle ground in the specificity-effort tradeoff curve. Text prompts, while easy to produce and capable of gallery-wide operations, struggle with spatial and structural specificity for fine-grained tasks \cite{koley2023you, chowdhury2022fs, chowdhury2023scenetrilogy, song2017fine, bhunia2023sketch2saliency}. For instance, precisely describing ``the zebra with distinctive stripes that curve inward at the shoulder'' through text alone presents significant challenges that sketches can naturally overcome. Interactive methods using clicks or scribbles, while offering precise spatial control, operate only on one image at a time and typically require multiple interactions \textit{per image} \cite{wong2024scribbleprompt}. This fundamental limitation makes them impractical for gallery-wide operations where users need to segment the same concept across multiple images simultaneously.

Sketches, by contrast, naturally combine semantic intent with structural specificity in a single interaction \cite{koley2023its}. This dual encoding enables our approach to segment visual concepts across entire image galleries through a single exemplar sketch -- a capability neither text nor interactive methods can match. The sketch serves as both a semantic concept descriptor (similar to text) and a structural template (similar to scribbles), making it uniquely suited for fine-grained segmentation tasks where spatial relationships and distinctive visual features play a crucial role.

Our approach confronts two significant technical challenges. First, we must operate without sketch-mask pairing data, as collecting such a dataset is infeasible given the sketch community's ongoing struggle to gather even basic sketch-photo pairs \cite{chowdhury2022fs, sangkloy2016the}. Second, we must bridge the substantial domain gap between abstract sketches and photos \cite{yelamarthi2018zero, sain2023clip} at the pixel level.

We address these challenges through a novel mask-free approach that strategically utilizes: \textit{(1)} a frozen Fine-Grained Sketch-Based Image Retrieval (FG-SBIR) model, and \textit{(2)} a pre-trained foundation model (CLIP \cite{radford2021learning} or DINOv2 \cite{oquab2023dinov2}). These components are linked by our central hypothesis: \textit{a sketch and its matching background-masked image will occupy proximate positions within a fixed FG-SBIR embedding space} \cite{sain2023clip}. This creates a self-supervised loop -- the better our segmentation mask highlights the relevant content, the closer the masked image and sketch will be in the embedding space, providing a natural training signal without manual annotations.

For mask generation, we extract spatial feature maps from photos and global feature vectors from sketches. Using cosine similarity, bilinear upscaling, and differentiable thresholding, we generate pseudo masks that highlight regions corresponding to the sketch query. To prevent trivial solutions and enable multi-granular segmentation, we introduce mask regularisation techniques and a novel sketch-partitioning strategy that enables part-level segmentation without additional training data.

Our primary contributions are:
\textit{(i)} A mask-free training paradigm that eliminates the need for pixel-perfect annotations while still achieving high-quality segmentation results.
\textit{(ii)} A synergistic framework that leverages sketch-based image retrieval models as critics to guide foundation models in generating accurate segmentation masks.
\textit{(iii)} Multi-granular segmentation capabilities that leverage sketches' unique structural encoding to precisely specify visual intent that text cannot effectively communicate -- from shape-specific distinctions between similar instances to exact part boundaries and spatial relationships between objects -- all within a single framework.
Empirical validation through extensive benchmark evaluations, demonstrating superior performance over existing approaches for both seen and unseen categories.

\vspace{-0.2cm}
\section{Related Works}
\vspace{-0.3cm}
\keypoint{Sketch for Vision Tasks.} \textit{Freehand sketches} has long established its significance for several visual understanding tasks \cite{bhunia2022doodle, yelamarthi2018zero, sain2023clip, chowdhury2023what, koley2023picture, bhunia2023sketch2saliency}. Following its success in $2D$ and $3D$ image retrieval \cite{yelamarthi2018zero, sain2023clip, dey2019doodle, luo2022structure, luo2020towards}, \textit{sketch} as a vision-modality has been used in image generation \cite{koley2023picture}, saliency detection \cite{bhunia2023sketch2saliency}, inpainting \cite{yu2019free}, shape-modelling \cite{mikaeili2023sked}, augmented reality \cite{luo2022structure}, medical image analysis \cite{kobayashi2023sketch}, image editing \cite{zeng2022sketchedit, lin2023sketchfacenerf}, object detection \cite{chowdhury2023what}, representation learning \cite{das2021sketchode}, class-incremental learning \cite{bhunia2022doodle}, etc.\cut{Despite its success in several vision tasks, using sketches for image segmentation is however still in its infancy.} To our best knowledge, Sketch-a-Segmenter \cite{hu2020sketch} is the only existing work dealing with sketch-guided segmentation.

\keypoint{Segmentation in Vision.} Image segmentation \cite{ronneberger2015u, chen2017deeplab, Chen_DeepLabv3+, long2015fully} can be broadly classified into three categories based on \textit{task granularity} -- \textit{(i) semantic} segmentation \cite{long2015fully,xie2021segformer,zheng2021rethinking} -- aims to classify and label each pixel into a set of pre-defined classes, \textit{(ii) instance} segmentation \cite{tian2020conditional, he2017mask} -- involves identifying different \textit{instances} of each object, \textit{(iii) panoptic} segmentation \cite{xu2023open, hwang2021exemplar} -- combines semantic and instance segmentation to simultaneously assign class labels and identify object instances. Rising popularity of vision-language \cite{radford2021learning} and generative \cite{rombach2022high, karras2019style} models further inspired \textit{open-vocab} segmentation frameworks like GroupViT \cite{xu2022groupvit}, SegCLIP \cite{luo2023segclip}, Segmentation-in-Style \cite{pakhomov2021segmentation}, ODISE \cite{xu2023open}, etc. The recent Segment Anything Model (SAM) \cite{kirillov2023segment} introduces a zero-shot framework capable of handling diverse segmentation tasks in different image distributions \cite{kirillov2023segment}. Its success promoted multiple follow-up works like HQ-SAM \cite{ke2023segment}, PerSAM \cite{zhang2023personalize},  etc. Our aim here is to introduce the subjectivity of freehand sketches for image segmentation. {Moreover, unlike prior methods, our approach segments \textit{all} similar instances of the query sketch from a large gallery of photos instead of segmenting a single image at hand.}

\keypoint{Weakly/Un-Supervised Segmentation.} Weakly supervised frameworks typically use image-level labels \cite{Araslanov_2020_CVPR, li2018tell, pathak2015constrained, pathak2014fully, pinheiro2015image}, scribbles \cite{lin2016scribblesup, vernaza2017learning}, bounding-boxes \cite{khoreva2017simple, papandreou2015weakly, dai2015boxsup}, cue points \cite{bearman2016s}, eye-tracking data \cite{papadopoulos2014training}, web-tags \cite{ahmed2014semantic}, saliency masks \cite{oh1701exploiting}, etc.\ as a source for weak-supervision. Such methods can be broadly classified as -- \textit{(i) one} stage and \textit{(ii) two} stage. While one-stage methods \cite{Araslanov_2020_CVPR, chen2020weakly,papandreou2015weakly, pathak2014fully} directly use image-level labels for training, they tend to perform lower than two-stage ones, which first generate pseudo segmentation labels via attention maps \cite{zhou2016learning, selvaraju2022grad} and then use them for training. In contrast, unsupervised image segmentation can be broadly divided into two categories-- \textit{(i) generative methods} utilise unlabelled images to train specialised image generators \cite{arandjelovic2019object, Chen_2019} or use pre-trained generators to directly generate segmentation masks \cite{MelasKyriazi_etal,Voynov_etal, abdal2021labels4free}; \textit{(ii) discriminative methods} are mainly based on clustering and contrastive learning \cite{Hwang_etal, Xie_etal, van2021unsupervised, zhang2020self}.
\cut{MaskContrast \cite{van2021unsupervised}, the current state-of-the-art in unsupervised semantic segmentation uses saliency detection to find object segments (\ie, foreground) and then learns pixel-wise embeddings over a contrastive objective.}

\vspace{-0.1cm}
\section{Revisiting Sketch-based Image Retrieval}
\label{sec:sbir}
Here we revisit the baseline SBIR paradigms in favour of the broader vision community. Baseline \textit{category-level} \cite{collomosse2019livesketch} and \textit{fine-grained} \cite{sain2023clip} SBIR models described here will be used later (\cref{sec:sketch_guided_mask}) to guide the training of our \textit{mask-free} sketch-based \textit{category-level} and \textit{fine-grained} image segmentation models respectively.

\keypoint{Baseline Category-level SBIR.} Provided a query sketch $\mathcal{S}\in\mathbb{R}^{H\times W\times 3}$ of any class, category-level SBIR aims to retrieve a photo $p_i^j$ from the \textit{same} class, out of a gallery $\mathcal{G}=\{p_i^j\}_{i=1}^{N_j}{|}_{j=1}^{N_c}$ containing images from a total $N_c$ classes with $N_j$ images per class \cite{sain2023clip}. Particularly, a backbone feature extractor (\textit{separate} weights for sketch and photo branches) network is trained to generate $d$-dimensional feature vector $f_i = \mathcal{F}(\mathcal{I}):\mathbb{R}^{H\times W\times 3}\rightarrow \mathbb{R}^{d}$ via triplet loss \cite{yu2016sketch} that aims to minimise the distance {$\delta(a,b)={||a-b||}_2$} between an anchor sketch ($\mathcal{S}$) feature $f_s$ and a positive photo ($\mathcal{P}$) feature $f_p$ from the same category as $\mathcal{S}$, while increasing that from a negative photo ($\mathcal{N}$) feature $f_n$ of a different category. With margin $\mu_{\mathrm{cat}}>0$, triplet loss for category-level SBIR could be formulated as: $\mathcal{L}_\mathrm{triplet}^{\mathrm{cat}} = \mathtt{max}\{0,\mu_{\mathrm{cat}}+\delta(f_s,f_p)-\delta(f_s,f_n)\}$

\keypoint{{Baseline Fine-grained (FG) SBIR}.} Unlike category-level, \textit{cross-category FG-SBIR} setup intends to learn a single model capable of fine-grained instance-level matching from multiple ($N_c$) classes. Akin to category-level \cite{sain2023clip}, cross-category FG-SBIR framework learns a backbone feature extractor $\mathcal{F}(\cdot)$
(weights \textit{shared} between sketch and photo branches), via triplet loss $\mathcal{L}^{\mathrm{fine}}_{\mathrm{triplet}}$ (similar to $\mathcal{L}^{\mathrm{cat}}_{\mathrm{triplet}}$) but with \textit{hard} triplets, where the negative sample is a \textit{different instance} ($\mathcal{P}^j_k;k\neq i$) of the \textit{same class} as the anchor sketch ($\mathcal{S}_i^j$) and its matching photo ($\mathcal{P}_i^j$) \cite{sain2023clip}. Furthermore, for learning inter-class discrimination, it uses an $N_c$-class classification head with cross-entropy loss ($\mathcal{L}_{\mathrm{class}}$) on the sketch-photo joint embedding space. Thus, the total loss \cite{bhunia2022adaptive} for baseline FG-SBIR is: $\mathcal{L}^{\mathrm{fine}}_{\mathrm{triplet}}+\mathcal{L}_{\mathrm{class}}$.

\begin{figure}[!htbp]
\vspace{-0.3cm}
    \centering
    \includegraphics[width=0.9\linewidth]{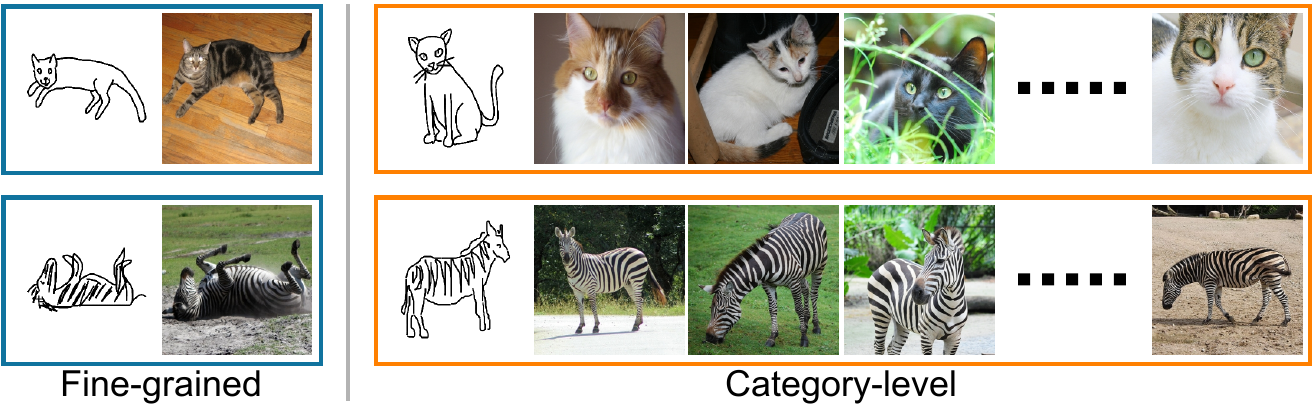}
    \vspace{-0.2cm}
    \caption{Examples of \textit{fine-grained} (left) and \textit{category-level} (right) image pairs from Sketchy \cite{sangkloy2016the} dataset.}
    \vspace{-0.4cm}
    \label{fig:datsets}
\end{figure}
\keypoint{Datasets.\ } Category-level datasets like Sketchy-Extended \cite{sangkloy2016the, liu2017deep}, TU-Berlin \cite{eitz2012humans}, Quick!, Draw \cite{ha2018neural}, etc.\ hold category-level association, where \textit{every} sketch of a category is associated with \textit{all} photos of the \textit{same} category as $\{s_i^j , \{ p_k^j\}_{k=1}^{N_j^P}\}_{i=1}^{N_j^S}|_{j=1}^{N_c}$ from $N_c$ classes with every $j^{th}$ class having $N^S_j$ sketches and $N^P_j$ photos. Whereas, FG-SBIR datasets (\eg, Sketchy \cite{sangkloy2016the}) contain instance-level sketch-photo pairs as $\{s_i^j,p_i^j\}_{i=1}^{N_j}{|}_{j=1}^{N_c}$ from $N_c$ classes with every $j^{th}$ class having $N_j$ sketch-photo pairs with \textit{fine-grained association}. Here we use category-level and fine-grained SBIR models \cite{sain2021stylemeup} trained with Sketchy-Extended \cite{sangkloy2016the} and Sketchy \cite{sangkloy2016the} datasets respectively.

\vspace{-0.1cm}
\section{Proposed Methodology}
\vspace{-0.1cm}
\keypoint{Overview.} Given an image $\mathcal{I}\in\mathbb{R}^{h\times w\times 3}$, state-of-the-art supervised image segmentation models \cite{long2015fully, chen2017deeplab, Chen_DeepLabv3+} aim to generate a segmentation mask $\mathcal{M}\in\mathbb{R}^{h\times w\times (c+1)}$ via pixel-level classification across a pre-defined set of $c$ classes and a background ($+1$). On the contrary, given a query sketch $\mathcal{S}$, our method generates a binary segmentation mask $\mathcal{M}\in\mathbb{R}^{h\times w\times 1}$, depicting the pixel-locations, wherever the \textit{queried-concept} appears in \textit{any} candidate image $\mathcal{I}\in\mathbb{R}^{h\times w\times 3}$, with pixels belonging to that concept marked as $1$ and all the rest as $0$. Learning segmentation frameworks via typical supervised training setup however, is impractical given the laborious process of collecting human-annotated ground truth segmentation masks. We therefore aim to utilise the existing sketch-photo dataset \cite{sangkloy2016the} to devise a weakly-supervised mask-free image segmenter. Although existing weak supervision signals (query) like tags \cite{ahmed2014semantic}, caption \cite{mukhoti2023open}, cue points \cite{bearman2016s}, etc.\ might suffice for category-level vision tasks, they lack in fine-grained cues unlike sketches \cite{bhunia2023sketch2saliency, chowdhury2023what}. Inspired by \textit{sketch}'s success in multiple fine-grained vision tasks (\eg, retrieval \cite{sain2023clip}, detection \cite{chowdhury2023what}, generation \cite{koley2023picture}, etc.), we aim to establish \textit{sketch} as a worthy query-modality for \textit{mask-free} weakly-supervised image segmentation, at \textit{category-level}, \textit{fine-grained}, and \textit{part-level} segmentation setup.

\vspace{-0.1cm}
\subsection{Problem Definition: Segmentation Granularity}
\vspace{-0.2cm}
\keypoint{Category Level Segmentation.}
In this setup, we aim to predict a binary segmentation mask, denoted as $\mathcal{M}\in \mathbb{R}^{h\times w \times 1}$, based on a query sketch $\mathcal{S}_c$ for a specific category, denoted as $c$. $\mathcal{M}$ will distinguish the foreground object (of category $c$) by marking the pixels inside the boundary as $1$, while the background pixels are set to $0$. Essentially, when you draw a ``camel'' this system will generate segmentation masks for any ``camel'' photos in the \textit{entire} test gallery.

\begin{figure}[!htbp]
\vspace{-0.3cm}
    \centering
    \includegraphics[width=0.85\columnwidth]{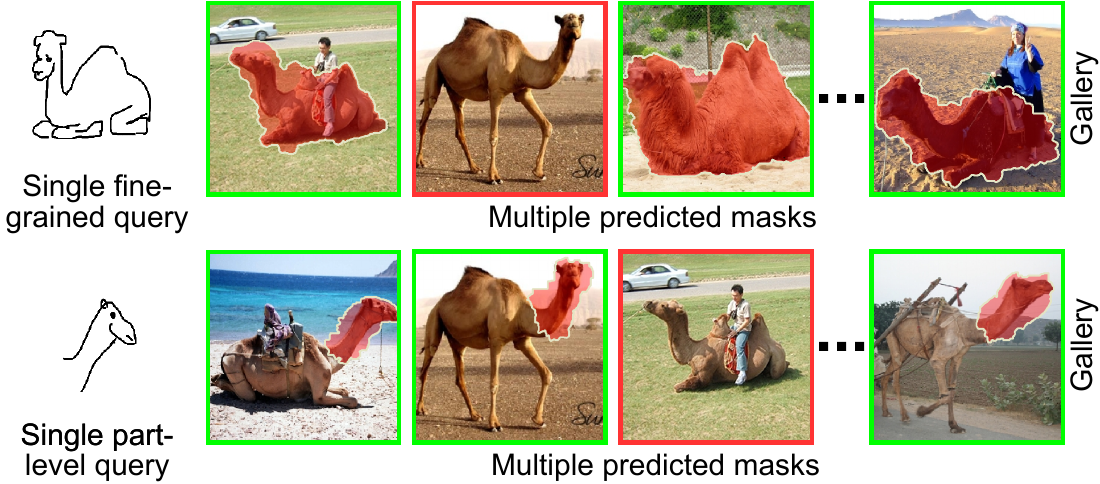}
    \vspace{-0.3cm}
    \caption{\textit{Fine-grained} (left) and \textit{part-level} (right) segmentation.}
    \vspace{-0.3cm}
    \label{fig:task}
\end{figure}

\keypoint{Fine-grained Segmentation.\ } Given a query sketch $\mathcal{S}_f$ depicting a specific \textit{object instance} of category $c$, we generate similar segmentation masks $\mathcal{M}\in \mathbb{R}^{h\times w \times 1}$ for any photo in the gallery of category $c$ where the object holds the \textit{same shape/pose} as that of $\mathcal{S}_f$. For instance, drawing a ``camel \textit{sitting down}'' should generate masks \textit{only} for those photos where the camel(s) is(are) ``sitting down'' in the \textit{same} shape/pose, not \textit{any} photo containing ``camel'' (\cref{fig:task} (left)).

\keypoint{Part-level Segmentation.} Unlike existing weak supervision signals (\eg, tags \cite{ahmed2014semantic}, cue points \cite{bearman2016s}, etc.), sketch has the ability to model \textit{part-level query} \cite{hung2019scops}. Accordingly, given a query sketch $\mathcal{S}_p$ depicting a \textit{specific part} of an object from category $c$, the aim is to generate similar segmentation masks $\mathcal{M}\in \mathbb{R}^{h\times w \times 1}$ depicting \textit{only the queried part} for any photo in $\mathcal{G}$ of category $c$. For instance, drawing ``a camel's head'' should generate segmentation masks showing only the ``head'' for \textit{any} ``camel'' photos in $\mathcal{G}$ (\cref{fig:task} (right)). Here the part-level query $\mathcal{S}_p$ is a region-wise \textit{spatial-subset} ($\mathcal{S}_p \subset \mathcal{S}$) of the full query $\mathcal{S}$.

\vspace{-0.1cm}
\subsection{Baseline Weakly Supervised Segmentation}
\label{sec:baseline_wss}

We consider the available sketches ($\mathcal{S}$) and photos ($\mathcal{P}$) from $\mathcal{D}_{\mathrm{cat}}$=$ \{\mathcal{S}_i,\mathcal{P}_i\}_{i=1}^{M}$ and $\mathcal{D}_{\mathrm{fine}}$=$\{\mathcal{S}_j,\mathcal{P}_j\}_{j=1}^{N}$ with category-level and fine-grained association between them respectively. We design the baseline sketch-based image segmenter taking inspiration from text-based weakly-supervised segmentation method \cite{mukhoti2023open} that uses patch-wise InfoNCE loss between CLIP's \textit{global textual feature} and \textit{patch-wise visual features} \cite{mukhoti2023open} for better text-photo \textit{region-wise} alignment (instead of using \textit{global} feature for both as in CLIP training) required for dense pixel-level segmentation task. We however exclude the textual encoder and employ CLIP's visual encoder to extract \textit{patch-wise} photo features and \textit{global} query sketch feature to apply the same loss.

In particular, we consider a pre-trained CLIP \cite{radford2021learning} ViT-L/14 backbone visual feature extractor $\mathbf{V}(\cdot)$ that splits the input $\mathcal{I}$ into $T$ patches and extracts patch-wise features $\{\mathcal{I}_p^i\}$ and global $\mathtt{[CLS]}$-token feature $\mathcal{I}_\mathtt{CLS}$ as $\mathbf{V}(\mathcal{I})$ = $[\mathcal{I}_p^1,\mathcal{I}_p^2,\cdots,\mathcal{I}_p^T,\mathcal{I}_\mathtt{CLS}]\in\mathbb{R}^{(T+1)\times d}$. For a sketch-query $\mathcal{S}$, we use its \textit{global} feature representation $\mathbf{s}=\mathbf{V}_g(\mathcal{S})\in\mathbb{R}^{1\times d}$ using the $\mathtt{[CLS]}$-token feature, while for a candidate photo $\mathcal{P}$, we utilise its \textit{patch-wise} feature representation $\mathbf{p}=\mathbf{V}_p(\mathcal{P})\in\mathbb{R}^{T\times d}$. Although $\mathbf{V}$ is \textit{shared} between sketch and photo, $\mathbf{V}_g$ and $\mathbf{V}_p$  denotes \textbf{g}lobal and \textbf{p}atch-wise feature extraction respectively. We calculate cosine similarity between $\mathbf{s}$ and $\mathbf{p}$ to obtain a patch-wise correlation vector $\xi \in \mathbb{R}^{T\times 1}$ with $\mathtt{SoftMax}$ normalisation across the patch dimension. Finally, taking the weighted sum across the patches, we obtain the sketch-query aware photo feature as ${\mathbf{\Hat{p}}} = \sum_{i=1}^{T}(\xi_i\times \mathbf{p}_i)$. We perform $l_2$ normalisation on $\mathbf{s}$ and $\mathbf{\Hat{p}}$ followed by calculating InfoNCE between them as:

\vspace{-0.4cm}
\begin{notsotiny}
\begin{equation}
    \mathcal{L}_{\mathrm{InfoNCE}} = \frac{1}{2k}\sum_{i=1}^{k}\left(\frac{\mathtt{exp}({\mathbf{\Hat{p}}_i\cdot\mathbf{s}_i})}{\sum_{j=1}^{k}\mathtt{exp}({\mathbf{\Hat{p}}_i\cdot\mathbf{s}_j})}+\frac{\mathtt{exp}({\mathbf{\Hat{p}}_i\cdot\mathbf{s}_i})}{\sum_{j=1}^{k}\mathtt{exp}({\mathbf{\Hat{p}}_j\cdot\mathbf{s}_i})}\right)
    \vspace{-0.1cm}
\end{equation}
\end{notsotiny}

\begin{figure}[!t]
\vspace{-0.3cm}
    \centering
    \includegraphics[width=1\columnwidth]{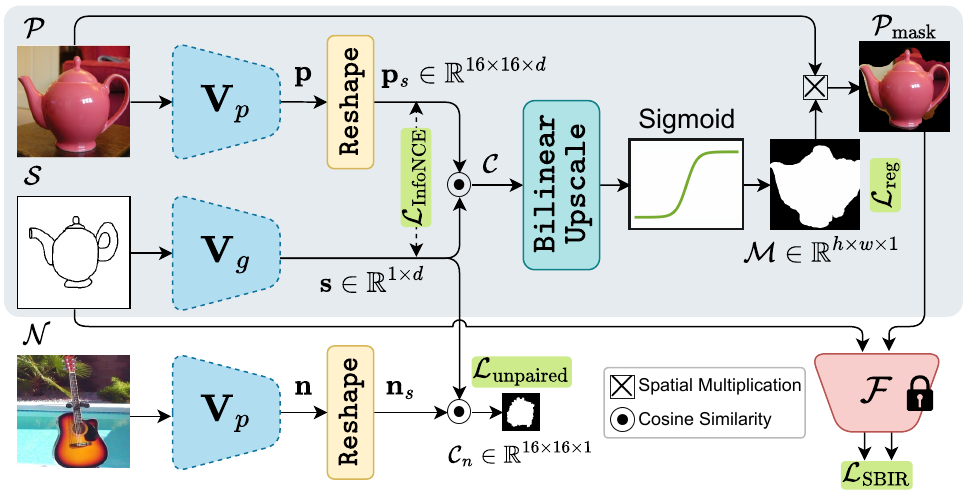}
    \vspace{-0.7cm}
    \caption{We generate sketch-guided correlation map $\mathcal{C}$ by multiplying the \textit{global} sketch feature $\mathbf{s}$ with reshaped \textit{patch-wise} photo feature $\mathbf{p}_s$. Bilinear upscaling, followed by \textit{differentiable} Sigmoid thresholding yields patch-wise correlation mask $\mathcal{M}$, which when multiplied with $\mathcal{P}$ gives \textit{masked candidate photo} $\mathcal{P}_{\mathrm{mask}}$ (highlighted in \colorbox{gray}{grey}). Frozen SBIR backbone $\mathcal{F}$ enforces the input sketches to generate masks that segment \textit{only} the foreground object (queried via $\mathcal{S}$) via $\mathcal{L}_{\mathrm{SBIR}}$. $\mathcal{L}_{\mathrm{infoNCE}}$ between $\mathbf{s}$ and $\mathbf{p}_s$ enhances sketch-photo alignment. At every batch, $\mathcal{L}_{\mathrm{unpaired}}$ ensures an \textit{all-zero} mask for negative samples ($\mathcal{N}$). To avoid overfitting to a trivial \textit{all-one} mask, we additionally regularise $\mathcal{M}$ via $\mathcal{L}_{\mathrm{reg}}$.}
    \vspace{-0.4cm}
    \label{fig:arch}
\end{figure}
\noindent Here, $k$ denotes batchsize.\ For a photo of size $\mathbb{R}^{224\times 224\times 3}$, the number of patches $T$=$ 256$ for ViT-L/14 backbone. During testing, we calculate cosine similarity between \textit{global} sketch query feature $\mathbf{s}$ $\in$ $\mathbb{R}^{1\times d}$ and \textit{patch-wise} candidate image features $\mathbf{p}$ $\in$ $\mathbb{R}^{256\times d}$ to obtain the patch-wise correlation vector $\xi$ $\in$ $\mathbb{R}^{256\times 1}$ that shows the degree to which each image patch correlates to the query-concept. We apply element-wise $\mathtt{Sigmoid}$ on $\xi$ and reshape its spatial tensor equivalent of size $\mathbb{R}^{16\times 16\times 1}$ followed by bilinearly upscaling it to the original photo size, to get the similarity map of size $\mathbb{R}^{h\times w\times 1}$.\ Empirically thresholding it with a fixed value ($0.5$) we get the predicted binary segmentation mask.

However, this baseline suffers a few drawbacks -- \textit{(i)} while CLIP \cite{radford2021learning} works reasonably well on category-level tasks \cite{sain2023clip}, it has limited fine-grained potential \cite{sain2023clip}. Consequently, for the fine-grained setup, we use a pre-trained DINOv2 \cite{oquab2023dinov2} ViT-L/14 backbone which showcases better fine-grained cross-modal correspondence \cite{oquab2023dinov2}, and \textit{(ii)} CLIP's contrastive learning happens in the \textit{embedding-space} \cite{radford2021learning}, while segmentation is a \textit{pixel-level} task. Furthermore, alignment in the latent embedding-space does not guarantee pixel-level semantic alignment. While this embedding-space alignment works well in case of \textit{semantic-driven} text-based segmentation \cite{mukhoti2023open}, sketches being more \textit{shape-driven} \cite{sain2023clip, bhunia2020sketch, chowdhury2023scenetrilogy, chowdhury2022fs} perform sub-optimally in this case.

\vspace{-0.1cm}
\subsection{Sketch-guided Mask Generation}
\vspace{-0.1cm}
\label{sec:sketch_guided_mask}
Given a sketch and a candidate image, we extract the \textit{global} sketch-query feature $\mathbf{s}$=$\mathbf{V}_g(\mathcal{S})\in\mathbb{R}^{1\times d}$ and \textit{patch-wise} candidate image features $\mathbf{p}$=$\mathbf{V}_p(\mathcal{P})\in\mathbb{R}^{T\times d}$. We reshape (for brevity) the \textit{patch-wise} features $\mathbf{p}$ in the equivalent image-space spatial order to get the $\mathbf{p}_s$=$\mathtt{Reshape}(\mathbf{p})\in\mathbb{R}^{h'\times w'\times d}$ spatial tensor where $T$=$h'\times w'$. We get the \textit{sketch-guided correlation map} $\mathcal{C}$ by computing cosine similarity between $\mathbf{s}$ and $\mathbf{p}_s$ using the sketch as a \textit{probe-vector} as:

\vspace{-0.45cm}
\begin{small}
\begin{equation}
    \mathcal{C}(u,v) = \frac{\langle\mathbf{s},\mathbf{p}_s(u,v)\rangle}{||\mathbf{s}||~||\mathbf{p}_s(u,v)||}~~~~u=[1, .., h'], v=[1, .., w']
    \vspace{-0.1cm}
    \label{eq:corr_map}
\end{equation}
\end{small}

Now the research question remains as to how can we leverage this sketch-guided correlation map $\mathcal{C}$ in the learning process while backpropagating the training signal via pixel space. One solution could be to bilinearly upscale $\mathcal{C}\in\mathbb{R}^{h'\times w'\times 1}$ to the original photo resolution ($\mathbb{R}^{h\times w\times 1}$), and threshold it with a fixed value and multiply with the candidate photo to get a \textit{masked photo}, where ideally the foreground objects (w.r.t.\ the sketch concept) would remain \textit{unmasked} and the background regions would be \textit{suppressed}. Nonetheless, hard thresholding with a fixed value invokes \textit{non-differentiability} due to the involvement of a Heaviside function \cite{chen2021localizing}. To alleviate this issue, we approximate the Heaviside thresholding \cite{chen2021localizing} operation with a temperature ($\tau$) controlled Sigmoid function \cite{patel2022recall, chen2021localizing} as:

\vspace{-0.3cm}
\begin{equation}
    \mathcal{M}={1} \big/ {(1+\mathtt{exp}((\mathcal{C}-0.5)/\tau))}
    \vspace{-0.15cm}
\end{equation}

\noindent where $\tau$ is the temperature parameter defining the sharpness of the Sigmoid, and $0.5$ is the thresholding value. With the thresholded patch-wise correlation mask ${\mathcal{M}}$, we generate the masked candidate photo {via spatial-wise multiplication} as $\mathcal{P}_{\mathrm{mask}}$=$\mathcal{P}\times{\mathcal{M}}$. During training, we aim to jointly optimise the visual encoder such that the mask generates \textit{true} responses, \textit{only} in those spatial regions of the \textit{photo} that contain the object drawn in the input \textit{sketch}. Now, to quantify the quality of the masked photo in terms of \textit{background-foreground separation}, we use a pre-trained SBIR \cite{sain2021stylemeup} model-based loss described in the next section.

\vspace{-0.1cm}
\subsection{Training Objectives}
\vspace{-0.1cm}
\keypoint{Pre-trained SBIR Loss.} We hypothesise that the masked photo $\mathcal{P}_{\mathrm{mask}}$ and the query sketch $\mathcal{S}$ will lie very close in the pre-trained SBIR model's ($\mathcal{F}$) embedding space when the generated mask $\mathcal{M}$ would \textit{only} segment the foreground object related to the sketch query. However, when the generated mask falsely suppresses the foreground object, $\mathcal{P}_{\mathrm{mask}}$ would likely diverge from $\mathcal{S}$, in the same space. Therefore, $\mathcal{F}$ is used as a \textit{critique} that would provide a training signal to $\mathbf{V}$ to generate accurate foreground masks that would place $\mathcal{P}_{\mathrm{mask}}$ and $\mathcal{S}$ in close proximity in the latent space; otherwise a high loss is imposed. Given a distance function $\delta(\cdot,\cdot)$ the SBIR loss becomes:

\vspace{-0.4cm}
\begin{equation}
    \mathcal{L}_{\mathrm{SBIR}} = \delta(\mathcal{F}(\mathcal{P}_{\mathrm{mask}}),\mathcal{F}(\mathcal{S}))
    \vspace{-0.1cm}
\end{equation}

Notably, for category-level and fine-grained segmentation, we use pre-trained category-level and fine-grained SBIR models \cite{sain2021stylemeup} respectively.

\keypoint{Unpaired Sketch-Photo Loss.} Apart from paired sketch-photo data, we focus on \textit{unpaired} photos to ensure that \textit{nothing} is segmented (correlation map contains only $0$s) from them, as they do \textit{not} contain the sketched query concept. We implement this via unpaired sketch-photo loss, where, across each batch, for every query-sketch $\mathcal{S}$, we sample a random non-paired image $\mathcal{N}$ (from different instances for fine-grained, and different classes for category-level). We compute and reshape the \textit{patch-wise} features of $\mathcal{N}$ as $\mathbf{n}_s$=$\mathbf{V}_p(\mathcal{N})\in\mathbb{R}^{h'\times w'\times d}$ and calculate cosine similarity between $\mathbf{n}_s$ and $\mathbf{s}$, followed by $\mathtt{Sigmoid}$ normalisation to generate the correlation map $\mathcal{C}_n$. The model is constrained to predict an \textit{all-zero} $\mathcal{C}_n$ for random non-paired candidate images via a binary cross-entropy loss as:

\vspace{-0.5cm}
\begin{equation}
    \mathcal{L}_{\mathrm{unpaired}} = -\frac{1}{h'w'}\sum_{i=1}^{h'}\sum_{j=1}^{w'}\mathtt{log}(1-\mathcal{C}_n(i,j)))
    \vspace{-0.3cm}
\end{equation}

\keypoint{Mask Regularisation Loss.}
Ideally, we want to segment \textit{only} the foreground sketch query concept suppressing the background pixels. Nonetheless, minimising $\mathcal{L}_{\mathrm{SBIR}}$ might result in a trivial solution of classifying \textit{all} pixels as foreground object \cite{zeng2019multi}. To eliminate the chance of a trivial mask with \textit{all} $1$s, we utilise a regularisation loss that computes the cross-entropy between the generated mask $\mathcal{M}\in\mathbb{R}^{h\times w\times 1}$ and an all-zero mask as:

\vspace{-0.2cm}
\begin{equation}
    \mathcal{L}_{\mathrm{reg}} = -\frac{1}{hw}\sum_{i=1}^{h}\sum_{j=1}^{w}\mathtt{log}(1-\mathcal{M}(i,j)))
    \vspace{-0.15cm}
\end{equation}

Thus, our overall training objective (\cref{fig:arch}) becomes $\mathcal{L}_{\mathrm{total}}=\lambda_1 \mathcal{L}_{\mathrm{InfoNCE}}+\lambda_2 \mathcal{L}_{\mathrm{SBIR}} + \lambda_3 \mathcal{L}_{\mathrm{unpaired}} + \lambda_4 \mathcal{L}_{\mathrm{reg}}$.

\subsection{Further Extension to Part-level Segmentation}
\label{sec:aug}
\begin{figure}[!htbp]
\vspace{-0.2cm}
    \centering
    \includegraphics[width=0.95\columnwidth]{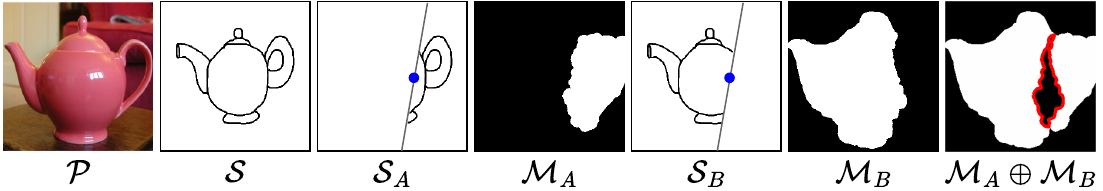}
    \vspace{-0.3cm}
    \caption{Example of our sketch-partitioning augmentation. $\mathcal{S}$ is divided into $\mathcal{S}_A$ and $\mathcal{S}_B$ based on the straight lines from centroid (\blue{blue} dot). Common foreground region of $\mathcal{M}$ is bordered in \red{red}.}
    \vspace{-0.3cm}
    \label{fig:aug}
\end{figure}

\vspace{-0.1cm}
Part-level segmentation aims to restrict the model to segment \textit{only} the region corresponding to the part-level sketch query (\eg, ``head'' of a camel). Unfortunately, common sketch-photo datasets \cite{sangkloy2016the} neither hold part-level sketches, nor part-level masks. We therefore aim to model this part-level behaviour via a simple sketch augmentation trick where we synthetically split the input sketch into multiple non-overlapping portions. Specifically, we first calculate the centroid of the query sketch from its vector coordinates. Now, at every pass, we draw a straight line from the centroid with a \textit{random} slope $m_1$. Based on the number of non-overlapping portions $n$, we calculate the slope values ($m_2, m_3, \cdots, m_n$) for the next lines as $m_{n}$=$m_{n-1}+\frac{360}{n}$. These $n$ straight lines segment the sketch into $n$ orthogonal non-overlapping parts (\cref{fig:aug}).
With $n$=$2$ (decided empirically), we segment the query sketch $\mathcal{S}$ into two non-overlapping parts $\mathcal{S}_A$ and $\mathcal{S}_B$. Using sketch-guided mask generation procedure (\cref{sec:sketch_guided_mask}), we generate the corresponding masks $\mathcal{M}_A$ and $\mathcal{M}_B$. Ideally, $\mathcal{M}_A$ and $\mathcal{M}_B$ should only segment their corresponding regions from the candidate images in a \textit{non-overlapping} manner, which therefore incurs a penalisation if $\mathcal{M}_A$ and $\mathcal{M}_B$ overlap. To do so we combine $\mathcal{M}_A$ and $\mathcal{M}_B$ via bit-wise XOR to get the final mask as $\mathcal{M}$=$\mathcal{M}_A \oplus \mathcal{M}_B$, where the \textit{common foreground regions} should be \textit{nullified}. Consequently, the same regions will be suppressed in the masked candidate image $\mathcal{P}_{\mathrm{mask}}$=$\mathcal{P}\times{\mathcal{M}}$ as well. Accordingly, $\mathcal{L}_{\mathrm{SBIR}}$ would impose a high loss, leading to better part-specific segmentation.

\vspace{-0.1cm}
\subsection{Discussion}
\vspace{-0.1cm}
For category-level segmentation, we use a pre-trained category-level SBIR model (\cref{sec:sbir}) with a pre-trained CLIP \cite{radford2021learning} ViT-L/14 backbone due to its category-level generalisation potential \cite{radford2021learning}. Whereas, for fine-grained segmentation, we use a pre-trained FG-SBIR model (\cref{sec:sbir}) with a pre-trained DINOv2 \cite{oquab2023dinov2} ViT-L/14 backbone. For extension to part-level, we use the same fine-grained segmentation model with the sketch-partitioning augmentation (\cref{sec:aug}) applied to $50\%$ of sketch samples in each batch.

During testing, $\mathbf{V}_g$ and $\mathbf{V}_p$ extract global sketch feature $\mathbf{s}$ and patch-wise candidate image features $\mathbf{p}_s$, using which we calculate the sketch-guided correlation map $\mathcal{C}$ (\cref{eq:corr_map}). Upscaling $\mathcal{C}$, followed by hard thresholding it with a fixed value of $0.5$ yields the final segmentation mask $\mathcal{M}$, which then undergoes standard CRF \cite{krahenbuhl2011efficient}-based post-processing.

\begin{table*}[!htbp]
\vspace{-7mm}
    \centering
    \setlength{\tabcolsep}{9.5pt}
    \renewcommand{\arraystretch}{0.8}
    \setlength{\aboverulesep}{1.5pt}
    \setlength{\belowrulesep}{1.5pt}
    \scriptsize
    \begin{tabular}{clcccc|cccc}
    \toprule
        \multicolumn{2}{c}{\multirow{2}{*}{\textbf{Methods}}} & \multicolumn{4}{c|}{\multirow{1}{*}{\textbf{Category-level}}} & \multicolumn{4}{c}{\multirow{1}{*}{\textbf{Fine-grained}}}\\\cmidrule(lr){3-6}\cmidrule(lr){7-10}
        & & {\textbf{mIoU (S)}} & \textbf{mIoU (U)} & \textbf{hIoU} & \textbf{pAcc.} & {\textbf{mIoU (S)}} & \textbf{mIoU (U)} & \textbf{hIoU} & \textbf{pAcc.}\\\cmidrule(lr){1-6}\cmidrule(lr){7-10}
       \multirow{4}{*}{Baselines} & {S-B1}          & 60.5 & 51.8 & 54.7 & 61.5 & 50.7 & 41.8 & 42.5 & 57.8\\
        & {S-B2}          & 67.7 & 55.2 & \bf60.4 & 69.7 & 51.3 & 46.1 & 47.6 & 61.6\\
        & {S-B3}          & 56.2 & 47.6 & 51.2 & 58.8 & 49.1 & 40.9 & 42.7 & 50.5\\
        & {T-B1}          & 66.3 & 56.7 & 57.9 & 70.2 & 50.2 & 42.7 & 43.2 & 61.7\\\cmidrule(lr){1-6}\cmidrule(lr){7-10}
        \multirow{4}{*}{Adapted SOTAs} & S-PerSAM          & 67.7 & 47.6 & 53.2 & 69.9 & 54.2 & 40.6 & 47.1 & 60.6\\
        & S-ZS-Seg          & 68.2 & 56.9 & 54.5 & 71.1 & 56.4 & 48.8 & 49.6 & 67.4\\
        & S-Graph-Cut       & 56.9 & 46.2 & 50.4 & 56.6 & 44.1 & 36.7 & 39.8 & 47.2\\

        \rowcolor{MidnightBlue!30}
        \multicolumn{2}{c}{\textbf{Ours}}    & \bf70.1 & \bf61.3 & 59.7 & \bf76.8  & \bf66.9 & \bf56.2 & \bf52.8 & \bf74.7\\

        \bottomrule
    \end{tabular}
    \vspace{-0.3cm}
    \caption{Quantitative comparison on the Sketchy-Extended \cite{liu2017deep} and Sketchy \cite{sangkloy2016the} for category-level and fine-grained segmentation.}

    \label{tab:mainTable}
    \vspace{-0.3cm}
\end{table*}

\begin{figure*}[!htbp]
    \centering
    \includegraphics[width=0.9\linewidth]{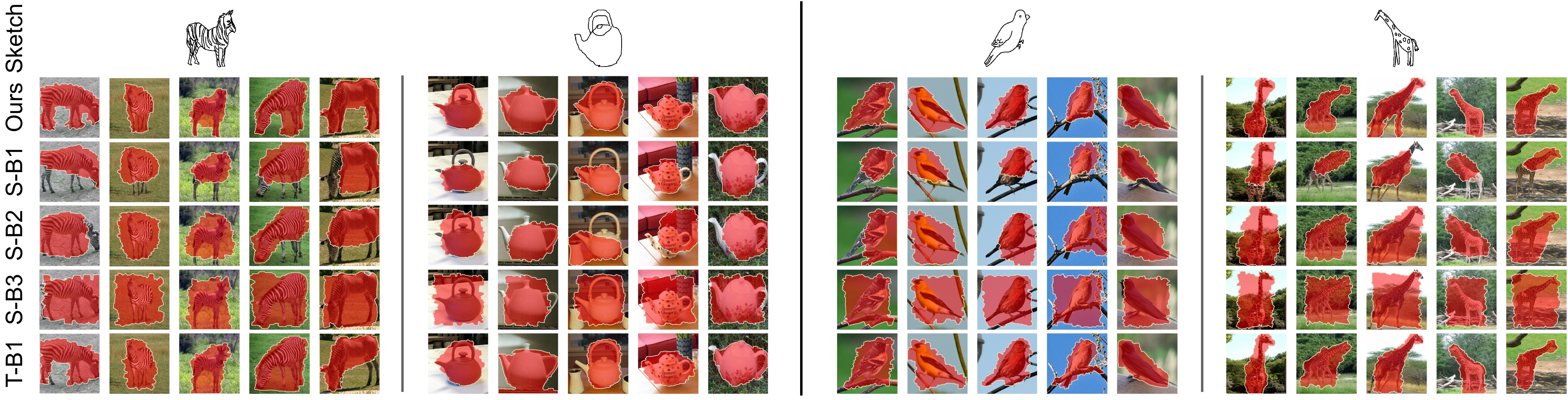}
    \vspace{-0.3cm}
    \caption{Qualitative comparison on Sketchy-Extended \cite{liu2017deep} for category-level segmentation on seen (left) and unseen (right) classes.}
    \label{fig:category}
    \vspace{-0.3cm}
\end{figure*}

\begin{figure*}[!htbp]
    \centering
    \includegraphics[width=0.9\linewidth]{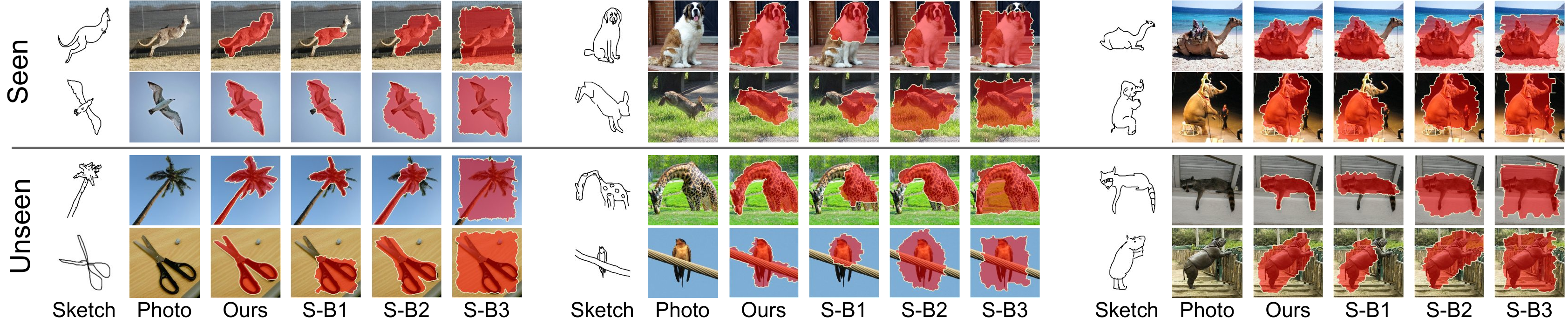}
    \vspace{-0.3cm}
    \caption{Qualitative comparison on Sketchy \cite{sangkloy2016the} for fine-grained segmentation on seen (top) and unseen (bottom) classes.}
    \label{fig:finegrained}
    \vspace{-0.5cm}
\end{figure*}

\vspace{-0.2cm}
\section{Experiments}
\vspace{-0.2cm}
\keypoint{Dataset.\ } As our method \textit{does not} require ground truth masks, we can utilise any standard sketch-photo dataset for training. Here, we use Sketchy \cite{sangkloy2016the} and Sketchy-extended \cite{liu2017deep, sangkloy2016the} to train our category-level and fine-grained models respectively. While Sketchy \cite{sangkloy2016the} holds $12,500$ photos from $125$ classes, each having at least $5$ fine-grained paired sketches, its extended version \cite{liu2017deep} carries additional $60,652$ photos from ImageNet \cite{deng2009imagenet}.

\keypoint{Evaluation.\ } We evaluate for both category-level (Sketchy-extended) and fine-grained (Sketchy) setups individually on two paradigms -- \textit{(i)} for \textit{unseen set} evaluation, we train our method on $104$ Sketchy classes and test on the $21$ unseen classes; \textit{(ii)} for \textit{seen set} evaluation, {we randomly divide the $104$ seen-classes into $80$:$20$ train:test splits across samples.} To evaluate segmentation accuracy, we manually annotate each test image from the Sketchy dataset with segmentation label masks to form the $\mathtt{SketchySegment}$ dataset. For further research, we open-sourced the $\mathtt{SketchySegment}$ dataset (\textit{cf}.\ \red{\S} Supplementary). Notably, our training pipeline \textit{does not} involve any ground truth masks. Following standard literature \cite{zhou2023zegclip, xu2022simple, mukhoti2023open}, we use mean intersection over union (mIoU), and per-pixel classification accuracy (pAcc.) as our primary evaluation metrics for both seen and unseen sets. Additionally, we also report the harmonic mean IoU (hIoU) among the unseen and seen sets.

\keypoint{Implementation Details.}
We use a pre-trained ViT-L/14 CLIP \cite{radford2021learning} and ViT-L/14 DINOv2 \cite{oquab2023dinov2} backbone for category-level and fine-grained segmentation models respectively. We fine-tune the $\mathtt{LayerNorm}$ layers of the CLIP visual encoder with a learning rate of $10^{-5}$, keeping the rest of the parameters frozen. The DINOv2 encoder is fine-tuned with a low learning rate of $10^{-6}$. We train the model for $200$ epochs using AdamW \cite{loshchilov2019decoupled} optimiser and a batch size of $16$. All $\lambda$ values are set to $1$, empirically. We use pre-trained baseline SBIR and FG-SBIR models from \cite{sain2021stylemeup}.

\label{sec:competitors}
\keypoint{Competitors.} Due to the unavailability of sketch-based image segmentation frameworks, we compare our method with a few self-designed baselines and some naively adopted SOTAs \cite{zhang2023personalize, xu2022simple}. \textit{(i) \textbf{{\underline{S}}}ketch-based Baselines:} $\bullet$~\textbf{S-B1} is the same as our baseline weakly supervised segmentation method described in \cref{sec:baseline_wss}.\cut{This naive baseline is nonetheless sub-optimal due to its embedding-space alignment.} $\bullet$ \textbf{S-B2} omits the unpaired Sketch-Photo Loss $\mathcal{L}_{\mathrm{unpaired}}$, keeping the rest of the model same as ours. $\bullet$ \textbf{S-B3} excludes the mask regularisation loss $\mathcal{L}_{\mathrm{reg}}$, keeping the rest of the model intact.\cut{In other words, this baseline depicts the importance of mask regularisation in avoiding the trivial solution of producing an \textit{all-one mask}.} \textit{(ii) \textbf{{\underline{T}}}ext-based Baseline:} $\bullet$ \textbf{T-B1} To compare the efficacy of our sketch-based framework with conventional text-based segmenter, we design \textbf{T-B1}, following architecture of SegCLIP \cite{luo2023segclip} where during inference, we additionally feed the class-label as a handcrafted prompt \cite{zhou2022learning} $\mathtt{``a~photo~of~a~[CLASS]"}$.
\textit{(iii) \textbf{{\underline{S}}}ketch-adopted SOTAs:} In \textbf{S-PerSAM}, given a sketch-photo pair ($\mathcal{S, P}$), we first calculate a pseudo mask $\xi$ using our baseline segmenter (\cref{sec:baseline_wss}). Then, with the ($\xi, \mathcal{P}$) pair, we use the pre-trained PerSAM \cite{zhang2023personalize} model to generate the background and foreground prompts. These prompts and candidate images upon passing through pre-trained SAM \cite{kirillov2023segment} yield the final segmentation masks.
\textbf{S-ZS-Seg} naively adapts the two-stage text-based zero-shot segmentation framework of \cite{xu2022simple} for sketches by replacing its CLIP text encoder with a CLIP vision encoder \cite{radford2021learning} to encode input sketches. For \textbf{S-Graph-Cut}, we first generate pseudo segmentation masks using Graph-Cut \cite{boykov2006graph} by using input \textit{sketch-strokes} as the \textit{foreground-marker}, which are then used to train a supervised CNN model \cite{long2015fully} to generate final segmentation masks. \textbf{S-Sketch-a-Segmenter} \cite{hu2020sketch} trains a DeepLabv3$+$-based fully-supervised segmentation model \cite{Chen_DeepLabv3+} and a sketch-hypernetwork \cite{hu2020sketch} with \textit{fully-annotated pixel-level masks} for each Sketchy \cite{sangkloy2016the} dataset images.

\vspace{-0.1cm}
\subsection{Performance Analysis}
\vspace{-0.2cm}

\keypoint{Category-level Segmentation.} \cref{tab:mainTable} shows the following observations -- \textit{(i)} although \textbf{S-B2} outperforms \textbf{S-B1} (due to sub-optimal embedding-space alignment) and \textbf{S-B3} (owing to lack of mask regularisation), all these baselines fare poorly when compared to ours, particularly for \textbf{S-B3} ($19.82\%$ mIoU (S) drop \vs ours), justifying the necessity of mask regularisation via $\mathcal{L}_{\mathrm{reg}}$. \textit{(ii)} \textbf{S-ZS-Seg} with its explicit zero-shot modelling \cite{xu2022simple} depicts comparable performance on both seen and unseen classes. However, \textbf{S-PerSAM} despite leveraging the large-scale pre-training of SAM \cite{kirillov2023segment}, turns out to be sub-optimal for the unseen classes. \textit{(iii)} owing to the inferior pseudo-mask generation, \textbf{S-Graph-Cut} fails to beat even the stronger baselines (\ie, \textbf{S-B2}, \textbf{T-B1}). Our method outperforms these competitors with an average $8.63~(14.49)\%$ mIoU gain on the seen (unseen) classes, even without any ground truth masks during training. Qualitative comparisons are shown in \cref{fig:category}.

\keypoint{Fine-grained Segmentation.} Fine-grained setup in \cref{tab:mainTable} shows: \textit{(i)} although \textbf{T-B1} competes ours in category-level setup, it scores a lower mIoU in the more challenging fine-grained setup. We posit that, while simple handcrafted prompts (\ie, $\mathtt{a~photo~of~a~[CLASS]}$) of \textbf{T-B1} sufficed the category-level task, it failed to capture the fine-grained shape/structure. {Although adding more descriptions might improve performance for \textbf{T-B1}, the research question we raise is: \textit{how fine-grained a textual description can be?} \textit{Fine-grained} captions generated by SOTA captioners \cite{li2022blip, li2023blip} often suffer from noise and inaccuracy \cite{li2022blip, li2023blip}. Additionally, crafting detailed textual prompts is more laborious compared to freehand sketching \cite{bhunia2022doodle, bhunia2023sketch2saliency, chowdhury2023what}.} \textit{(ii)} despite being a stronger category-level baseline, \textbf{S-B2} performs sub-optimally ($23.31\%$ mIoU (S) drop \vs ours) here, verifying the need for $\mathcal{L}_{\mathrm{unpaired}}$ in fine-grained setting. \textit{(iii)} our method outperforms all sketch-adapted SOTAs with an average hIoU margin of $13.84\%$ without the complicated positive-negative prior calculation of \textbf{S-PerSAM}, time-consuming two-stage approach of \textbf{S-ZS-Seg}, or the allegedly unstable pseudo-mask generation of \textbf{S-Graph-Cut}. Qualitative comparisons (\cref{fig:finegrained}) show the fine-grained generalisation potential of our method. \textit{(iv)} \textbf{S-Sketch-a-Segmenter} despite being \textit{fully-supervised}, achieves lower hIoU and pAcc.\ in the Sketchy dataset.

\begin{figure}[!htbp]
    \centering
    \includegraphics[width=1\columnwidth]{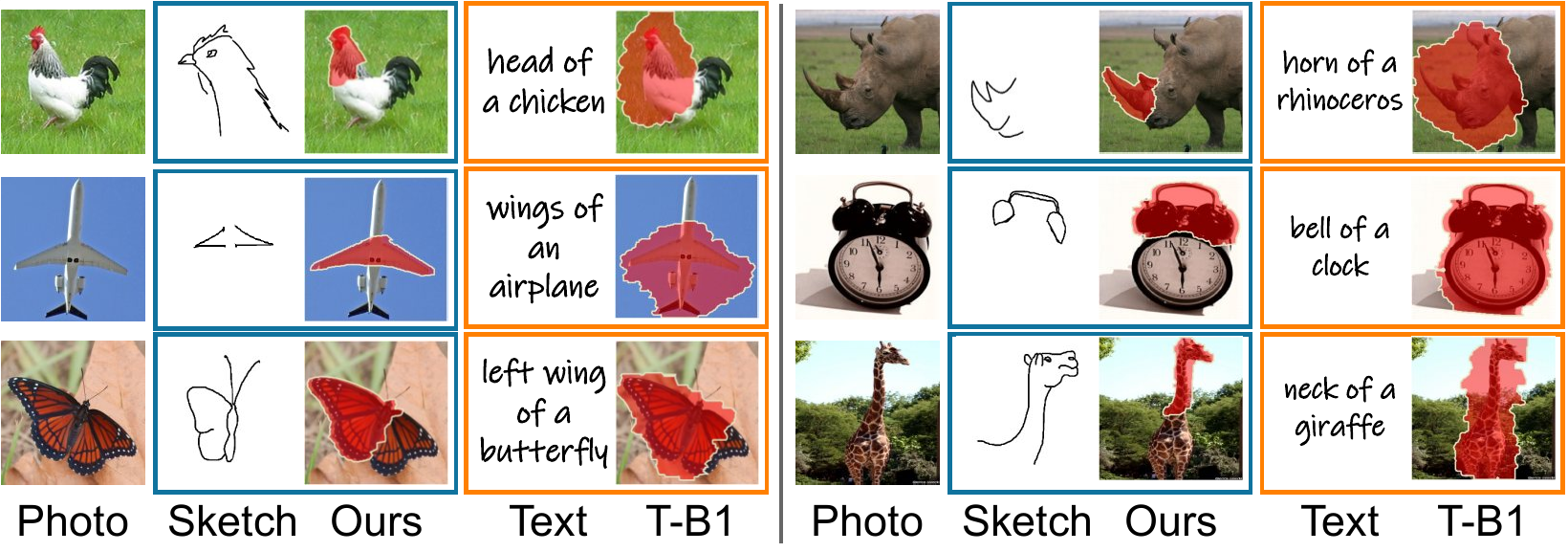}
    \vspace{-0.7cm}
    \caption{Part-level segmentation: \textbf{T-B1} \vs Ours on Sketchy \cite{sangkloy2016the}.}
    \label{fig:part}
    \vspace{-0.5cm}
\end{figure}

\keypoint{Part-level Segmentation.} Delving deeper to judge our method's potential for part-level segmentation (\eg, only the ``head'' of a camel), we find quantitative assessment to be infeasible, as part-level ground truth annotations are unavailable. We thus assess the part-level segmentation capability of our method qualitatively, with a set of manually edited sketches depicting a certain part of an object instance (\eg, ``wing'' of an ``airplane''). \cref{fig:part} shows a few images comparing the proposed method with \textbf{T-B1}. While \textbf{T-B1} often produces imprecise segmentation masks due to lingual-ambiguity of text-query \cite{schwartz2023discriminative}, our method equipped with sketch-partitioning augmentation produces part-level segmentation masks that accurately captures the \textit{semantic-intent} of end-user provided via query sketch. Furthermore, here we calculate \textit{Mean Opinion Score (MOS)} \cite{huynh2010study} by asking $10$ users to draw $25$ part-level sketches and rate on a scale of $1$ to $5$ (bad$\shortrightarrow$excellent) based on their opinion of segmentation quality. Accordingly, we obtain a higher MOS ($\mu$$\pm$$\sigma$ of $250$ responses) of $4.01$$\pm$$0.7$ compared to $3.15$$\pm$$0.2$ and $3.09$$\pm$$0.4$ of \textbf{S-B1} and \textbf{T-B1} respectively.

\vspace{-0.2cm}

\begin{figure}[!htbp]
    \centering
    \includegraphics[width=1\columnwidth]{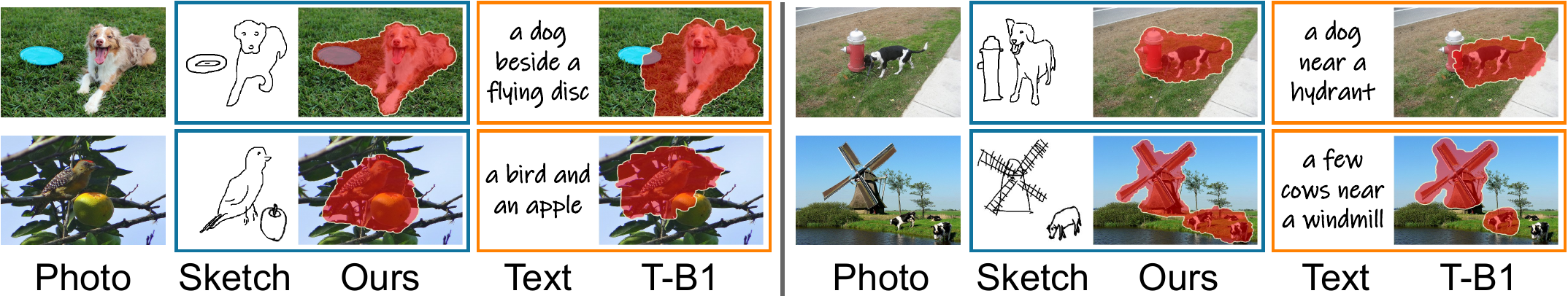}
    \vspace{-0.6cm}
    \caption{Composed segmentation: \textbf{T-B1} \vs \textbf{Ours}.}
    \label{fig:composed}
    \vspace{-0.4cm}
\end{figure}

\keypoint{Segmenting Composed Concepts.} We also test our framework for \textit{composed concept} segmentation. For instance, drawing a \textit{``dog'' sitting beside a ``frisbee''} should segment \textit{both} the dog and the frisbee. Qualitative comparison (on FS-COCO \cite{chowdhury2022fs}) in \cref{fig:composed} shows our method to perform reasonably better in segmenting composed concepts than \textbf{T-B1}.

\vspace{-0.1cm}

\vspace{-0.1cm}
\subsection{Ablation on Design}
\label{sec:abal}
\vspace{-0.2cm}
\keypoint{{[i]} Backbone Variants.} Performance of our sketch-based image segmenter depends crucially on the choice of feature extractor backbone. Accordingly, we explore with multiple publicly available pre-trained DINOv2 \cite{oquab2023dinov2} and CLIP \cite{radford2021learning} backbones. Quantitative results in \cref{tab:abal}, show DINOv2 ViT-L/14 and CLIP ViT-L/14 to achieve the highest mIoU for fine-grained and category-level segmentation respectively. Although CLIP \cite{radford2021learning} and DINOv2 \cite{oquab2023dinov2} showcase comparable category-level performance, we posit that, due to its self-supervised pre-training, DINOv2 \cite{oquab2023dinov2} outperforms CLIP \cite{radford2021learning} in the fine-grained setup.

\keypoint{{[ii]} Does Fine-tuning entire CLIP help?} Existing literature on sketch-based vision tasks \cite{sangkloy2022sketch, sain2023clip} point towards fine-tuning the \textit{entire} CLIP vision encoder to adapt it to their respective tasks \cite{sangkloy2022sketch, sain2023clip}. Doing so for training our category-level segmentation model however, significantly drops the final seen/unseen class mIoU (\textbf{CLIP Fine-tuning} result in \cref{tab:abal}). We posit that fine-tuning all parameters of the vision encoder distorts the pre-trained knowledge of CLIP, thus sacrificing its generalisation potential.

\keypoint{{[iii]} Effect of $\mathcal{L}_{\mathrm{InfoNCE}}$.} We posit that the InfoNCE loss aids in proper alignment of sketch-photo features to generate accurate patch-wise correlation maps. Hence, a drop of $12.69~(17.48)\%$ seen-class mIoU in \textbf{w/o $\mathcal{L}_{\mathrm{InfoNCE}}$}, for category-level (fine-grained) setup, verifies its importance in learning patch-wise sketch-photo region alignment.

\keypoint{{[iv]} Why Sketch-Partitioning Augmentation?\ } Quantitative assessment of sketch-partitioning for part-level segmentation is infeasible due to the lack of annotation masks. However \textbf{w/o Sketch-partitioning} results in \cref{tab:abal} shows our part-level augmentation to be particularly helpful for our fine-grained segmentation model, as its absence further reduces the seen (unseen) class mIoU by $8.22~(10.32)\%$.

\vspace{-0.3cm}
\begin{table}[!htbp]
\centering
\setlength{\tabcolsep}{7pt}
\renewcommand{\arraystretch}{0.5}
\setlength{\aboverulesep}{1pt}
\setlength{\belowrulesep}{1pt}
\notsotiny
\label{tab:abal}
\begin{tabular}{llcc|cc}
\toprule
\multicolumn{2}{c}{\multirow{2}{*}{\textbf{Methods}}} & \multicolumn{2}{c|}{\textbf{Category-level}} & \multicolumn{2}{c}{\textbf{Fine-grained}}\\
\cmidrule(lr){3-4}\cmidrule(lr){5-6}
& & \textbf{mIoU (S)} & \textbf{mIoU (U)} & \textbf{mIoU (S)} & \textbf{mIoU (U)}\\\cmidrule(lr){1-4}\cmidrule(lr){5-6}
\multicolumn{1}{c}{\multirow{3}{*}{DINOv2}} &  ViT-S/14              & 58.7& 50.9 & 60.4  & 50.3\\
& ViT-B/14              & 65.1  & 57.2 & 63.6  & 51.7\\
& ViT-L/14              & 68.5  & 59.8 & \bf66.9  & \bf56.2\\\cmidrule(lr){1-6}
\multicolumn{1}{c}{\multirow{3}{*}{CLIP}} & ViT-B/16                & 59.7  & 51.2 & 54.6  & 43.3\\
& ViT-B/32                & 66.4  & 59.9 & 56.4 & 44.8\\
& ViT-L/14                & \bf70.1  & \bf61.3 & 60.1  & 49.7\\\cmidrule(lr){1-6}
\multicolumn{2}{c}{w/o $\mathcal{L}_{\text{InfoNCE}}$} & 61.2  & 52.8 & 55.2  & 48.1\\
\multicolumn{2}{c}{w/o Sketch-partitioning} & 69.3  & 60.1 & 61.4  & 50.4\\
\multicolumn{2}{c}{CLIP Fine-tuning}  & 60.6  & 32.7 & --  & --\\
\rowcolor{MidnightBlue!30}
\multicolumn{2}{c}{\textbf{\textit{Ours-full}}}                 & \bf70.1  & \bf61.3 & \bf66.9  & \bf56.2\\

\bottomrule
\end{tabular}
\vspace{-0.2cm}
\caption{Ablation on design.}
\vspace{-0.5cm}
\end{table}

\subsection{Limitations and Failure Cases}
\vspace{-0.2cm}

\begin{figure}[!htbp]
    \centering
    \includegraphics[width=1\columnwidth]{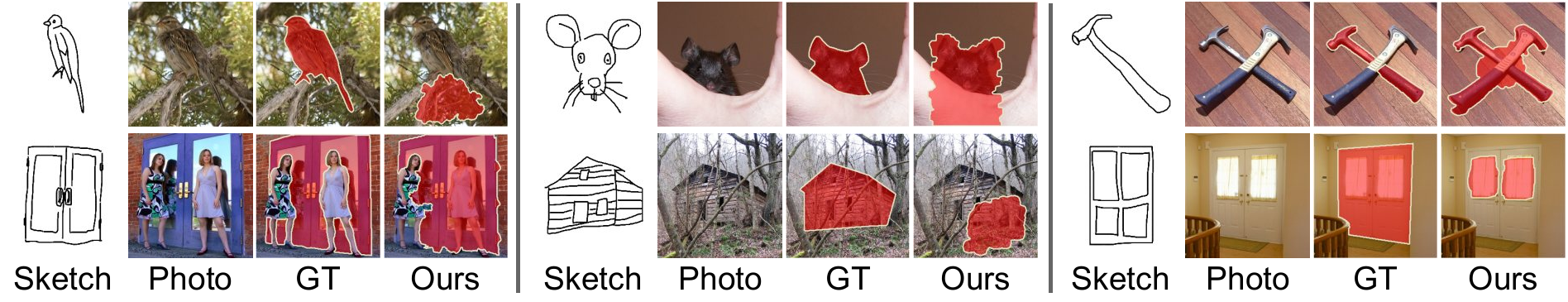}
    \vspace{-0.7cm}
    \caption{Failure cases. \textit{(Best view when zoomed in.)}}
    \label{fig:failure}
    \vspace{-0.3cm}
\end{figure}

Although our method is quite robust against deformed sketches and hard-to-segment candidate images (\cref{fig:category}-\ref{fig:composed}), it sometimes struggles to yield accurate segmentation masks in case of challenging candidate images (\cref{fig:failure}) with reflection, cluttered background, occlusion, or camouflage.

\vspace{-0.3cm}

\vspace{-0.3cm}
\section{Conclusion}
\vspace{-0.3cm}
Here, we explore the potential of freehand abstract sketches for \textit{subjective} image segmentation. Leveraging large-scale pre-trained visual encoders and SBIR models, we devise a \textit{mask-free} image segmenter that allows for segmentation at multiple granularity levels (\ie, category-level and fine-grained). Furthermore, using the smart augmentation trick of sketch-partitioning, we extend our method to part-level segmentation. Extensive qualitative and quantitative results depict the efficacy of our method.

{
    \small
    \bibliographystyle{ieeenat_fullname}
    \bibliography{main}

\begin{thebibliography}{96}
\providecommand{\natexlab}[1]{#1}
\providecommand{\url}[1]{\texttt{#1}}
\expandafter\ifx\csname urlstyle\endcsname\relax
  \providecommand{\doi}[1]{doi: #1}\else
  \providecommand{\doi}{doi: \begingroup \urlstyle{rm}\Url}\fi

\bibitem[Abdal et~al.(2021)Abdal, Zhu, Mitra, and Wonka]{abdal2021labels4free}
Rameen Abdal, Peihao Zhu, Niloy~J Mitra, and Peter Wonka.
\newblock {Labels4Free: Unsupervised Segmentation Using StyleGAN}.
\newblock In \emph{CVPR}, 2021.

\bibitem[Ahmed et~al.(2014)Ahmed, Cohen, and Price]{ahmed2014semantic}
Ejaz Ahmed, Scott Cohen, and Brian Price.
\newblock {Semantic Object Selection}.
\newblock In \emph{CVPR}, 2014.

\bibitem[Arandjelovic and Zisserman(2019)]{arandjelovic2019object}
Relja Arandjelovic and Andrew Zisserman.
\newblock {Object Discovery with a Copy-Pasting GAN}.
\newblock \emph{arXiv preprint arXiv:1905.11369}, 2019.

\bibitem[Araslanov and Roth(2020)]{Araslanov_2020_CVPR}
Nikita Araslanov and Stefan Roth.
\newblock {Single-Stage Semantic Segmentation From Image Labels}.
\newblock In \emph{CVPR}, 2020.

\bibitem[Bearman et~al.(2016)Bearman, Russakovsky, Ferrari, and
  Fei-Fei]{bearman2016s}
Amy Bearman, Olga Russakovsky, Vittorio Ferrari, and Li Fei-Fei.
\newblock {What's the Point: Semantic Segmentation with Point Supervision}.
\newblock In \emph{ECCV}, 2016.

\bibitem[Bhunia et~al.(2020)Bhunia, Yang, Hospedales, Xiang, and
  Song]{bhunia2020sketch}
Ayan~Kumar Bhunia, Yongxin Yang, Timothy~M Hospedales, Tao Xiang, and Yi-Zhe
  Song.
\newblock {Sketch Less for More: On-the-Fly Fine-Grained Sketch Based Image
  Retrieval}.
\newblock In \emph{CVPR}, 2020.

\bibitem[Bhunia et~al.(2022{\natexlab{a}})Bhunia, Gajjala, Koley, Kundu, Sain,
  Xiang, and Song]{bhunia2022doodle}
Ayan~Kumar Bhunia, Viswanatha~Reddy Gajjala, Subhadeep Koley, Rohit Kundu,
  Aneeshan Sain, Tao Xiang, and Yi-Zhe Song.
\newblock {Doodle It Yourself: Class Incremental Learning by Drawing a Few
  Sketches}.
\newblock In \emph{CVPR}, 2022{\natexlab{a}}.

\bibitem[Bhunia et~al.(2022{\natexlab{b}})Bhunia, Sain, Shah, Gupta, Chowdhury,
  Xiang, and Song]{bhunia2022adaptive}
Ayan~Kumar Bhunia, Aneeshan Sain, Parth Shah, Animesh Gupta, Pinaki~Nath
  Chowdhury, Tao Xiang, and Yi-Zhe Song.
\newblock {Adaptive Fine-Grained Sketch-Based Image Retrieval}.
\newblock In \emph{ECCV}, 2022{\natexlab{b}}.

\bibitem[Bhunia et~al.(2023)Bhunia, Koley, Kumar, Sain, Chowdhury, Xiang, and
  Song]{bhunia2023sketch2saliency}
Ayan~Kumar Bhunia, Subhadeep Koley, Amandeep Kumar, Aneeshan Sain, Pinaki~Nath
  Chowdhury, Tao Xiang, and Yi-Zhe Song.
\newblock {Sketch2Saliency: Learning to Detect Salient Objects from Human
  Drawings}.
\newblock In \emph{CVPR}, 2023.

\bibitem[Boykov and Funka-Lea(2006)]{boykov2006graph}
Yuri Boykov and Gareth Funka-Lea.
\newblock {Graph Cuts and Efficient N-D Image Segmentation}.
\newblock \emph{IJCV}, 2006.

\bibitem[Chen et~al.(2021{\natexlab{a}})Chen, Xie, Afouras, Nagrani, Vedaldi,
  and Zisserman]{chen2021localizing}
Honglie Chen, Weidi Xie, Triantafyllos Afouras, Arsha Nagrani, Andrea Vedaldi,
  and Andrew Zisserman.
\newblock {Localizing Visual Sounds the Hard Way}.
\newblock In \emph{CVPR}, 2021{\natexlab{a}}.

\bibitem[Chen et~al.(2020)Chen, Wu, Fu, Han, and Zhang]{chen2020weakly}
Liyi Chen, Weiwei Wu, Chenchen Fu, Xiao Han, and Yuntao Zhang.
\newblock {Weakly Supervised Semantic Segmentation with Boundary Exploration}.
\newblock In \emph{ECCV}, 2020.

\bibitem[Chen et~al.(2017)Chen, Papandreou, Kokkinos, Murphy, and
  Yuille]{chen2017deeplab}
Liang-Chieh Chen, George Papandreou, Iasonas Kokkinos, Kevin Murphy, and
  Alan~L. Yuille.
\newblock {DeepLab: Semantic Image Segmentation with Deep Convolutional Nets,
  Atrous Convolution, and Fully Connected CRFs}.
\newblock \emph{IEEE-TPAMI}, 2017.

\bibitem[Chen et~al.(2018)Chen, Zhu, Papandreou, Schroff, and
  Adam]{Chen_DeepLabv3+}
Liang-Chieh Chen, Yukun Zhu, George Papandreou, Florian Schroff, and Hartwig
  Adam.
\newblock {Encoder-Decoder with Atrous Separable Convolution for Semantic Image
  Segmentation}.
\newblock In \emph{ECCV}, 2018.

\bibitem[Chen et~al.(2019)Chen, Artières, and Denoyer]{Chen_2019}
Mickael Chen, Thierry Artières, and Ludovic Denoyer.
\newblock {Unsupervised Object Segmentation by Redrawing}.
\newblock In \emph{NeurIPS}, 2019.

\bibitem[Chen et~al.(2021{\natexlab{b}})Chen, Zhao, Yu, Zhang, and
  Duan]{chen2021conditional}
Xi Chen, Zhiyan Zhao, Feiwu Yu, Yilei Zhang, and Manni Duan.
\newblock {Conditional Diffusion for Interactive Segmentation}.
\newblock In \emph{ICCV}, 2021{\natexlab{b}}.

\bibitem[Chowdhury et~al.(2023{\natexlab{a}})Chowdhury, Bhunia, Sain, Koley,
  Xiang, and Song]{chowdhury2023scenetrilogy}
Pinaki~Nath Chowdhury, Ayan~Kumar Bhunia, Aneeshan Sain, Subhadeep Koley, Tao
  Xiang, and Yi-Zhe Song.
\newblock {SceneTrilogy: On Human Scene-Sketch and its Complementarity with
  Photo and Text}.
\newblock In \emph{CVPR}, 2023{\natexlab{a}}.

\bibitem[Chowdhury et~al.(2023{\natexlab{b}})Chowdhury, Bhunia, Sain, Koley,
  Xiang, and Song]{chowdhury2023what}
Pinaki~Nath Chowdhury, Ayan~Kumar Bhunia, Aneeshan Sain, Subhadeep Koley, Tao
  Xiang, and Yi-Zhe Song.
\newblock {What Can Human Sketches Do for Object Detection?}
\newblock In \emph{CVPR}, 2023{\natexlab{b}}.

\bibitem[Collomosse et~al.(2019)Collomosse, Bui, and
  Jin]{collomosse2019livesketch}
John Collomosse, Tu Bui, and Hailin Jin.
\newblock {LiveSketch: Query perturbations for guided sketch-based visual
  search}.
\newblock In \emph{CVPR}, 2019.

\bibitem[Dai et~al.(2015)Dai, He, and Sun]{dai2015boxsup}
Jifeng Dai, Kaiming He, and Jian Sun.
\newblock {BoxSup: Exploiting Bounding Boxes to Supervise Convolutional
  Networks for Semantic Segmentation}.
\newblock In \emph{ICCV}, 2015.

\bibitem[Das et~al.(2021)Das, Yang, Hospedales, Xiang, and
  Song]{das2021sketchode}
Ayan Das, Yongxin Yang, Timothy Hospedales, Tao Xiang, and Yi-Zhe Song.
\newblock {SketchODE: Learning neural sketch representation in continuous
  time}.
\newblock In \emph{ICLR}, 2021.

\bibitem[Deng et~al.(2009)Deng, Dong, Socher, Li, Li, and
  Fei-Fei]{deng2009imagenet}
Jia Deng, Wei Dong, Richard Socher, Li-Jia Li, Kai Li, and Li Fei-Fei.
\newblock Imagenet: A large-scale hierarchical image database.
\newblock In \emph{CVPR}, 2009.

\bibitem[Dey et~al.(2019)Dey, Riba, Dutta, Llados, and Song]{dey2019doodle}
Sounak Dey, Pau Riba, Anjan Dutta, Josep Llados, and Yi-Zhe Song.
\newblock {Doodle to Search: Practical Zero-Shot Sketch-based Image Retrieval}.
\newblock In \emph{CVPR}, 2019.

\bibitem[Eitz et~al.(2012)Eitz, Hays, and Alexa]{eitz2012humans}
Mathias Eitz, James Hays, and Marc Alexa.
\newblock {How do humans sketch objects?}
\newblock \emph{ACM TOG}, 2012.

\bibitem[Ha and Eck(2018)]{ha2018neural}
David Ha and Douglas Eck.
\newblock {A Neural Representation of Sketch Drawings}.
\newblock \emph{ICLR}, 2018.

\bibitem[Hao et~al.(2021)Hao, Liu, Wu, Han, Chen, Chen, Chu, Tang, Yu, Chen,
  et~al.]{hao2021edgeflow}
Yuying Hao, Yi Liu, Zewu Wu, Lin Han, Yizhou Chen, Guowei Chen, Lutao Chu,
  Shiyu Tang, Zhiliang Yu, Zeyu Chen, et~al.
\newblock {EdgeFlow: Achieving Practical Interactive Segmentation with
  Edge-Guided Flow}.
\newblock In \emph{ICCV}, 2021.

\bibitem[He et~al.(2017)He, Gkioxari, Dollar, and Girshick]{he2017mask}
Kaiming He, Georgia Gkioxari, Piotr Dollar, and Ross Girshick.
\newblock {Mask R-CNN}.
\newblock In \emph{ICCV}, 2017.

\bibitem[Hu et~al.(2020)Hu, Li, Yang, Hospedales, and Song]{hu2020sketch}
Conghui Hu, Da Li, Yongxin Yang, Timothy~M Hospedales, and Yi-Zhe Song.
\newblock {Sketch-a-Segmenter: Sketch-based Photo Segmenter Generation}.
\newblock \emph{IEEE TIP}, 2020.

\bibitem[Hung et~al.(2019)Hung, Jampani, Liu, Molchanov, Yang, and
  Kautz]{hung2019scops}
Wei-Chih Hung, Varun Jampani, Sifei Liu, Pavlo Molchanov, Ming-Hsuan Yang, and
  Jan Kautz.
\newblock {SCOPS: Self-Supervised Co-Part Segmentation}.
\newblock In \emph{CVPR}, 2019.

\bibitem[Huynh-Thu et~al.(2010)Huynh-Thu, Garcia, Speranza, Corriveau, and
  Raake]{huynh2010study}
Quan Huynh-Thu, Marie-Neige Garcia, Filippo Speranza, Philip Corriveau, and
  Alexander Raake.
\newblock {Study of Rating Scales for Subjective Quality Assessment of
  High-Definition Video}.
\newblock \emph{IEEE TBC}, 2010.

\bibitem[Hwang et~al.(2019)Hwang, Yu, Shi, Collins, Yang, Zhang, and
  Chen]{Hwang_etal}
J. Hwang, S. Yu, J. Shi, M. Collins, T. Yang, X. Zhang, and L. Chen.
\newblock {SegSort: Segmentation by Discriminative Sorting of Segments}.
\newblock In \emph{ICCV}, 2019.

\bibitem[Hwang et~al.(2021)Hwang, Oh, Lee, and Han]{hwang2021exemplar}
Jaedong Hwang, Seoung~Wug Oh, Joon-Young Lee, and Bohyung Han.
\newblock {Exemplar-Based Open-Set Panoptic Segmentation Network}.
\newblock In \emph{CVPR}, 2021.

\bibitem[Karras et~al.(2019)Karras, Laine, and Aila]{karras2019style}
Tero Karras, Samuli Laine, and Timo Aila.
\newblock {A Style-Based Generator Architecture for Generative Adversarial
  Networks}.
\newblock In \emph{CVPR}, 2019.

\bibitem[Ke et~al.(2023)Ke, Ye, Danelljan, Liu, Tai, Tang, and
  Yu]{ke2023segment}
Lei Ke, Mingqiao Ye, Martin Danelljan, Yifan Liu, Yu-Wing Tai, Chi-Keung Tang,
  and Fisher Yu.
\newblock {Segment Anything in High Quality}.
\newblock \emph{arXiv preprint arXiv:2306.01567}, 2023.

\bibitem[Khoreva et~al.(2017)Khoreva, Benenson, Hosang, Hein, and
  Schiele]{khoreva2017simple}
Anna Khoreva, Rodrigo Benenson, Jan Hosang, Matthias Hein, and Bernt Schiele.
\newblock {Simple Does It: Weakly Supervised Instance and Semantic
  Segmentation}.
\newblock In \emph{CVPR}, 2017.

\bibitem[Kirillov et~al.(2023)Kirillov, Mintun, Ravi, Mao, Rolland, Gustafson,
  Xiao, Whitehead, Berg, Lo, et~al.]{kirillov2023segment}
Alexander Kirillov, Eric Mintun, Nikhila Ravi, Hanzi Mao, Chloe Rolland, Laura
  Gustafson, Tete Xiao, Spencer Whitehead, Alexander~C Berg, Wan-Yen Lo, et~al.
\newblock {Segment Anything}.
\newblock In \emph{ICCV}, 2023.

\bibitem[Kobayashi et~al.(2023)Kobayashi, Gu, Hataya, Mizuno, Miyake, Watanabe,
  Takahashi, Takamizawa, Yoshida, Nakamura, et~al.]{kobayashi2023sketch}
Kazuma Kobayashi, Lin Gu, Ryuichiro Hataya, Takaaki Mizuno, Mototaka Miyake,
  Hirokazu Watanabe, Masamichi Takahashi, Yasuyuki Takamizawa, Yukihiro
  Yoshida, Satoshi Nakamura, et~al.
\newblock {Sketch-based Medical Image Retrieval}.
\newblock \emph{arXiv preprint arXiv:2303.03633}, 2023.

\bibitem[Koley et~al.(2023)Koley, Bhunia, Sain, Chowdhury, Xiang, and
  Song]{koley2023picture}
Subhadeep Koley, Ayan~Kumar Bhunia, Aneeshan Sain, Pinaki~Nath Chowdhury, Tao
  Xiang, and Yi-Zhe Song.
\newblock {Picture that Sketch: Photorealistic Image Generation from Abstract
  Sketches}.
\newblock In \emph{CVPR}, 2023.

\bibitem[Koley et~al.(2024{\natexlab{a}})Koley, Bhunia, Sain, Chowdhury, Xiang,
  and Song]{koley2023you}
Subhadeep Koley, Ayan~Kumar Bhunia, Aneeshan Sain, Pinaki~Nath Chowdhury, Tao
  Xiang, and Yi-Zhe Song.
\newblock {You'll Never Walk Alone: A Sketch and Text Duet for Fine-Grained
  Image Retrieval}.
\newblock In \emph{CVPR}, 2024{\natexlab{a}}.

\bibitem[Koley et~al.(2024{\natexlab{b}})Koley, Bhunia, Sekhri, Sain,
  Chowdhury, Xiang, and Song]{koley2023its}
Subhadeep Koley, Ayan~Kumar Bhunia, Deeptanshu Sekhri, Aneeshan Sain,
  Pinaki~Nath Chowdhury, Tao Xiang, and Yi-Zhe Song.
\newblock {It's All About Your Sketch: Democratising Sketch Control in
  Diffusion Models}.
\newblock In \emph{CVPR}, 2024{\natexlab{b}}.

\bibitem[Kr{\"a}henb{\"u}hl and Koltun(2011)]{krahenbuhl2011efficient}
Philipp Kr{\"a}henb{\"u}hl and Vladlen Koltun.
\newblock {Efficient Inference in Fully Connected CRFs with Gaussian Edge
  Potentials}.
\newblock In \emph{NeurIPS}, 2011.

\bibitem[Li et~al.(2022)Li, Li, Xiong, and Hoi]{li2022blip}
Junnan Li, Dongxu Li, Caiming Xiong, and Steven Hoi.
\newblock {BLIP: Bootstrapping language-image pre-training for unified
  vision-language understanding and generation}.
\newblock In \emph{ICML}, 2022.

\bibitem[Li et~al.(2023)Li, Li, Savarese, and Hoi]{li2023blip}
Junnan Li, Dongxu Li, Silvio Savarese, and Steven Hoi.
\newblock {BLIP-2: Bootstrapping Language-Image Pre-training with Frozen Image
  Encoders and Large Language Models}.
\newblock In \emph{ICML}, 2023.

\bibitem[Li et~al.(2018)Li, Wu, Peng, Ernst, and Fu]{li2018tell}
Kunpeng Li, Ziyan Wu, Kuan-Chuan Peng, Jan Ernst, and Yun Fu.
\newblock {Tell Me Where to Look: Guided Attention Inference Network}.
\newblock In \emph{CVPR}, 2018.

\bibitem[Lin et~al.(2016)Lin, Dai, Jia, He, and Sun]{lin2016scribblesup}
Di Lin, Jifeng Dai, Jiaya Jia, Kaiming He, and Jian Sun.
\newblock {ScribbleSup: Scribble-Supervised Convolutional Networks for Semantic
  Segmentation}.
\newblock In \emph{CVPR}, 2016.

\bibitem[Lin et~al.(2023)Lin, Feng-Lin, Shu-Yu, Kaiwen, Chunpeng, Lai, and
  Hongbo]{lin2023sketchfacenerf}
Gao Lin, Liu Feng-Lin, Chen Shu-Yu, Jiang Kaiwen, Li Chunpeng, Yukun Lai, and
  Fu Hongbo.
\newblock {SketchFaceNeRF: Sketch-based facial generation and editing in neural
  radiance fields}.
\newblock \emph{ACM TOG}, 2023.

\bibitem[Liu et~al.(2017)Liu, Shen, Shen, Liu, and Shao]{liu2017deep}
Li Liu, Fumin Shen, Yuming Shen, Xianglong Liu, and Ling Shao.
\newblock {Deep Sketch Hashing: Fast Free-hand Sketch-Based Image Retrieval}.
\newblock In \emph{CVPR}, 2017.

\bibitem[Long et~al.(2015)Long, Shelhamer, and Darrell]{long2015fully}
Jonathan Long, Evan Shelhamer, and Trevor Darrell.
\newblock {Fully Convolutional Networks for Semantic Segmentation}.
\newblock In \emph{CVPR}, 2015.

\bibitem[Loshchilov and Hutter(2019)]{loshchilov2019decoupled}
Ilya Loshchilov and Frank Hutter.
\newblock Decoupled weight decay regularization.
\newblock In \emph{ICLR}, 2019.

\bibitem[Luo et~al.(2023)Luo, Bao, Wu, He, and Li]{luo2023segclip}
Huaishao Luo, Junwei Bao, Youzheng Wu, Xiaodong He, and Tianrui Li.
\newblock {SegCLIP: Patch Aggregation with Learnable Centers for
  Open-Vocabulary Semantic Segmentation}.
\newblock In \emph{ICML}, 2023.

\bibitem[Luo et~al.(2020)Luo, Gryaditskaya, Yang, Xiang, and
  Song]{luo2020towards}
Ling Luo, Yulia Gryaditskaya, Yongxin Yang, Tao Xiang, and Yi-Zhe Song.
\newblock {Towards 3D VR-Sketch to 3D Shape Retrieval}.
\newblock In \emph{3DV}, 2020.

\bibitem[Luo et~al.(2022)Luo, Gryaditskaya, Xiang, and Song]{luo2022structure}
Ling Luo, Yulia Gryaditskaya, Tao Xiang, and Yi-Zhe Song.
\newblock {Structure-Aware 3D VR Sketch to 3D Shape Retrieval}.
\newblock In \emph{3DV}, 2022.

\bibitem[Melas-Kyriazi et~al.(2022)Melas-Kyriazi, Rupprecht, Laina, and
  Vedaldi]{MelasKyriazi_etal}
Luke Melas-Kyriazi, Christian Rupprecht, Iro Laina, and Andrea Vedaldi.
\newblock {Finding an Unsupervised Image Segmenter in each of your Deep
  Generative Models}.
\newblock In \emph{ICLR}, 2022.

\bibitem[Mikaeili et~al.(2023)Mikaeili, Perel, Cohen-Or, and
  Mahdavi-Amiri]{mikaeili2023sked}
Aryan Mikaeili, Or Perel, Daniel Cohen-Or, and Ali Mahdavi-Amiri.
\newblock {SKED: Sketch-guided Text-based 3D Editing}.
\newblock In \emph{CVPR}, 2023.

\bibitem[Mukhoti et~al.(2023)Mukhoti, Lin, Poursaeed, Wang, Shah, Torr, and
  Lim]{mukhoti2023open}
Jishnu Mukhoti, Tsung-Yu Lin, Omid Poursaeed, Rui Wang, Ashish Shah, Philip~HS
  Torr, and Ser-Nam Lim.
\newblock {Open Vocabulary Semantic Segmentation with Patch Aligned Contrastive
  Learning}.
\newblock In \emph{CVPR}, 2023.

\bibitem[Nath~Chowdhury et~al.(2022)Nath~Chowdhury, Sain, Gryaditskaya, Bhunia,
  Xiang, and Song]{chowdhury2022fs}
Pinaki Nath~Chowdhury, Aneeshan Sain, Yulia Gryaditskaya, Ayan~Kumar Bhunia,
  Tao Xiang, and Yi-Zhe Song.
\newblock {FS-COCO: Towards Understanding of Freehand Sketches of Common
  Objects in Context}.
\newblock In \emph{ECCV}, 2022.

\bibitem[Oh et~al.(2017)Oh, Benenson, Khoreva, Akata, Fritz, and
  Schiele]{oh1701exploiting}
Seong~Joon Oh, Rodrigo Benenson, Anna Khoreva, Zeynep Akata, Mario Fritz, and
  Bernt Schiele.
\newblock {Exploiting saliency for object segmentation from image level
  labels}.
\newblock In \emph{CVPR}, 2017.

\bibitem[Oquab et~al.(2023)Oquab, Darcet, Moutakanni, Vo, Szafraniec, Khalidov,
  Fernandez, Haziza, Massa, El-Nouby, et~al.]{oquab2023dinov2}
Maxime Oquab, Timoth{\'e}e Darcet, Th{\'e}o Moutakanni, Huy Vo, Marc
  Szafraniec, Vasil Khalidov, Pierre Fernandez, Daniel Haziza, Francisco Massa,
  Alaaeldin El-Nouby, et~al.
\newblock {DINOv2: Learning Robust Visual Features without Supervision}.
\newblock \emph{arXiv preprint arXiv:2304.07193}, 2023.

\bibitem[Pakhomov et~al.(2021)Pakhomov, Hira, Wagle, Green, and
  Navab]{pakhomov2021segmentation}
Daniil Pakhomov, Sanchit Hira, Narayani Wagle, Kemar~E Green, and Nassir Navab.
\newblock {Segmentation in Style: Unsupervised Semantic Image Segmentation with
  Stylegan and CLIP}.
\newblock \emph{arXiv preprint arXiv:2107.12518}, 2021.

\bibitem[Papadopoulos et~al.(2014)Papadopoulos, Clarke, Keller, and
  Ferrari]{papadopoulos2014training}
Dim~P Papadopoulos, Alasdair~DF Clarke, Frank Keller, and Vittorio Ferrari.
\newblock {Training Object Class Detectors from Eye Tracking Data}.
\newblock In \emph{ECCV}, 2014.

\bibitem[Papandreou et~al.(2015)Papandreou, Chen, Murphy, and
  Yuille]{papandreou2015weakly}
George Papandreou, Liang-Chieh Chen, Kevin~P Murphy, and Alan~L Yuille.
\newblock {Weakly- and Semi-Supervised Learning of a Deep Convolutional Network
  for Semantic Image Segmentation}.
\newblock In \emph{ICCV}, 2015.

\bibitem[Patel et~al.(2022)Patel, Tolias, and Matas]{patel2022recall}
Yash Patel, Giorgos Tolias, and Ji{\v{r}}{\'\i} Matas.
\newblock {Recall@k: Surrogate Loss with Large Batches and Similarity Mixup}.
\newblock In \emph{CVPR}, 2022.

\bibitem[Pathak et~al.(2015{\natexlab{a}})Pathak, Krahenbuhl, and
  Darrell]{pathak2015constrained}
Deepak Pathak, Philipp Krahenbuhl, and Trevor Darrell.
\newblock {Constrained Convolutional Neural Networks for Weakly Supervised
  Segmentation}.
\newblock In \emph{ICCV}, 2015{\natexlab{a}}.

\bibitem[Pathak et~al.(2015{\natexlab{b}})Pathak, Shelhamer, Long, and
  Darrell]{pathak2014fully}
Deepak Pathak, Evan Shelhamer, Jonathan Long, and Trevor Darrell.
\newblock {Fully Convolutional Multi-Class Multiple Instance Learning}.
\newblock In \emph{ICLR}, 2015{\natexlab{b}}.

\bibitem[Pinheiro and Collobert(2015)]{pinheiro2015image}
Pedro~O Pinheiro and Ronan Collobert.
\newblock {From Image-level to Pixel-level Labeling with Convolutional
  Networks}.
\newblock In \emph{CVPR}, 2015.

\bibitem[Radford et~al.(2021)Radford, Kim, Hallacy, Ramesh, Goh, Agarwal,
  Sastry, Askell, Mishkin, Clark, et~al.]{radford2021learning}
Alec Radford, Jong~Wook Kim, Chris Hallacy, Aditya Ramesh, Gabriel Goh,
  Sandhini Agarwal, Girish Sastry, Amanda Askell, Pamela Mishkin, Jack Clark,
  et~al.
\newblock {Learning Transferable Visual Models From Natural Language
  Supervision}.
\newblock In \emph{ICML}, 2021.

\bibitem[Rombach et~al.(2022)Rombach, Blattmann, Lorenz, Esser, and
  Ommer]{rombach2022high}
Robin Rombach, Andreas Blattmann, Dominik Lorenz, Patrick Esser, and Bj{\"o}rn
  Ommer.
\newblock {High-Resolution Image Synthesis with Latent Diffusion Models}.
\newblock In \emph{CVPR}, 2022.

\bibitem[Ronneberger et~al.(2015)Ronneberger, Fischer, and
  Brox]{ronneberger2015u}
Olaf Ronneberger, Philipp Fischer, and Thomas Brox.
\newblock {U-Net: Convolutional Networks for Biomedical Image Segmentation}.
\newblock In \emph{MICCAI}, 2015.

\bibitem[Sain et~al.(2021)Sain, Bhunia, Yang, Xiang, and
  Song]{sain2021stylemeup}
Aneeshan Sain, Ayan~Kumar Bhunia, Yongxin Yang, Tao Xiang, and Yi-Zhe Song.
\newblock {StyleMeUp: Towards Style-Agnostic Sketch-Based Image Retrieval}.
\newblock In \emph{CVPR}, 2021.

\bibitem[Sain et~al.(2023)Sain, Bhunia, Chowdhury, Koley, Xiang, and
  Song]{sain2023clip}
Aneeshan Sain, Ayan~Kumar Bhunia, Pinaki~Nath Chowdhury, Subhadeep Koley, Tao
  Xiang, and Yi-Zhe Song.
\newblock {CLIP for All Things Zero-Shot Sketch-Based Image Retrieval,
  Fine-Grained or Not}.
\newblock In \emph{CVPR}, 2023.

\bibitem[Sangkloy et~al.(2016)Sangkloy, Burnell, Ham, and
  Hays]{sangkloy2016the}
Patsorn Sangkloy, Nathan Burnell, Cusuh Ham, and James Hays.
\newblock {The Sketchy Database: Learning to Retrieve Badly Drawn Bunnies}.
\newblock \emph{ACM TOG}, 2016.

\bibitem[Sangkloy et~al.(2022)Sangkloy, Jitkrittum, Yang, and
  Hays]{sangkloy2022sketch}
Patsorn Sangkloy, Wittawat Jitkrittum, Diyi Yang, and James Hays.
\newblock {A Sketch Is Worth a Thousand Words: Image Retrieval with Text and
  Sketch}.
\newblock In \emph{ECCV}, 2022.

\bibitem[Schwartz et~al.(2023)Schwartz, Sn{\ae}bjarnarson, Chefer, Belongie,
  Wolf, and Benaim]{schwartz2023discriminative}
Idan Schwartz, V{\'e}steinn Sn{\ae}bjarnarson, Hila Chefer, Serge Belongie,
  Lior Wolf, and Sagie Benaim.
\newblock {Discriminative Class Tokens for Text-to-Image Diffusion Models}.
\newblock In \emph{ICCV}, 2023.

\bibitem[Selvaraju et~al.(2017)Selvaraju, Cogswell, Das, Vedantam, Parikh, and
  Batra]{selvaraju2022grad}
RR Selvaraju, M Cogswell, A Das, R Vedantam, D Parikh, and D Batra.
\newblock {Grad-CAM: Visual Explanations from Deep Networks via Gradient-based
  Localization}.
\newblock In \emph{ICCV}, 2017.

\bibitem[Song et~al.(2017)Song, Song, Xiang, and Hospedales]{song2017fine}
Jifei Song, Yi-Zhe Song, Tony Xiang, and Timothy~M Hospedales.
\newblock {Fine-Grained Image Retrieval: the Text/Sketch Input Dilemma.}
\newblock In \emph{BMVC}, 2017.

\bibitem[Tian et~al.(2020)Tian, Shen, and Chen]{tian2020conditional}
Zhi Tian, Chunhua Shen, and Hao Chen.
\newblock {Conditional Convolutions for Instance Segmentation}.
\newblock In \emph{ECCV}, 2020.

\bibitem[Van~Gansbeke et~al.(2021)Van~Gansbeke, Vandenhende, Georgoulis, and
  Van~Gool]{van2021unsupervised}
Wouter Van~Gansbeke, Simon Vandenhende, Stamatios Georgoulis, and Luc Van~Gool.
\newblock {Unsupervised Semantic Segmentation by Contrasting Object Mask
  Proposals}.
\newblock In \emph{ICCV}, 2021.

\bibitem[Vernaza and Chandraker(2017)]{vernaza2017learning}
Paul Vernaza and Manmohan Chandraker.
\newblock {Learning random-walk label propagation for weakly-supervised
  semantic segmentation}.
\newblock In \emph{CVPR}, 2017.

\bibitem[Voynov et~al.(2022)Voynov, Morozov, and Babenko]{Voynov_etal}
Andrey Voynov, Stanislav Morozov, and Artem Babenko.
\newblock {Object Segmentation Without Labels with Large-Scale Generative
  Models}.
\newblock In \emph{ICML}, 2022.

\bibitem[Wong et~al.(2024)Wong, Rakic, Guttag, and
  Dalca]{wong2024scribbleprompt}
Hallee~E Wong, Marianne Rakic, John Guttag, and Adrian~V Dalca.
\newblock {ScribblePrompt: Fast and Flexible Interactive Segmentation for Any
  Biomedical Image}.
\newblock In \emph{ECCV}, 2024.

\bibitem[Xie et~al.(2021)Xie, Wang, Yuille, Anandkumar, and
  Alvarez]{xie2021segformer}
Enze Xie, Wenhai Wang, Alan~L. Yuille, Anima Anandkumar, and Jose~M. Alvarez.
\newblock {SegFormer: Simple and Efficient Design for Semantic Segmentation
  with Transformers}.
\newblock In \emph{CVPR}, 2021.

\bibitem[Xie et~al.(2019)Xie, Liu, Hovy, Yuan, and Stoyanov]{Xie_etal}
Zhengyang Xie, Zhuang Liu, Eduard Hovy, Liangzhe Yuan, and Veselin Stoyanov.
\newblock {Invariant Information Clustering for Unsupervised Image
  Classification and Segmentation}.
\newblock In \emph{ICCV}, 2019.

\bibitem[Xu et~al.(2022{\natexlab{a}})Xu, De~Mello, Liu, Byeon, Breuel, Kautz,
  and Wang]{xu2022groupvit}
Jiarui Xu, Shalini De~Mello, Sifei Liu, Wonmin Byeon, Thomas Breuel, Jan Kautz,
  and Xiaolong Wang.
\newblock {GroupViT: Semantic Segmentation Emerges from Text Supervision}.
\newblock In \emph{CVPR}, 2022{\natexlab{a}}.

\bibitem[Xu et~al.(2023)Xu, Liu, Vahdat, Byeon, Wang, and De~Mello]{xu2023open}
Jiarui Xu, Sifei Liu, Arash Vahdat, Wonmin Byeon, Xiaolong Wang, and Shalini
  De~Mello.
\newblock {Open-Vocabulary Panoptic Segmentation with Text-to-Image Diffusion
  Models}.
\newblock In \emph{CVPR}, 2023.

\bibitem[Xu et~al.(2022{\natexlab{b}})Xu, Zhang, Wei, Lin, Cao, Hu, and
  Bai]{xu2022simple}
Mengde Xu, Zheng Zhang, Fangyun Wei, Yutong Lin, Yue Cao, Han Hu, and Xiang
  Bai.
\newblock {A Simple Baseline for Open-Vocabulary Semantic Segmentation with
  Pre-trained Vision-language Model}.
\newblock In \emph{ECCV}, 2022{\natexlab{b}}.

\bibitem[Yelamarthi et~al.(2018)Yelamarthi, Reddy, Mishra, and
  Mittal]{yelamarthi2018zero}
Sasi~Kiran Yelamarthi, Shiva~Krishna Reddy, Ashish Mishra, and Anurag Mittal.
\newblock {A Zero-Shot Framework for Sketch Based Image Retrieval}.
\newblock In \emph{ECCV}, 2018.

\bibitem[Yu et~al.(2019)Yu, Lin, Yang, Shen, Lu, and Huang]{yu2019free}
Jiahui Yu, Zhe Lin, Jimei Yang, Xiaohui Shen, Xin Lu, and Thomas~S Huang.
\newblock {Free-Form Image Inpainting with Gated Convolution}.
\newblock In \emph{ICCV}, 2019.

\bibitem[Yu et~al.(2016)Yu, Liu, Song, Xiang, Hospedales, and
  Loy]{yu2016sketch}
Qian Yu, Feng Liu, Yi-Zhe Song, Tao Xiang, Timothy~M Hospedales, and
  Chen-Change Loy.
\newblock {Sketch Me That Shoe}.
\newblock In \emph{CVPR}, 2016.

\bibitem[Zeng et~al.(2019)Zeng, Zhuge, Lu, Zhang, Qian, and Yu]{zeng2019multi}
Yu Zeng, Yunzhi Zhuge, Huchuan Lu, Lihe Zhang, Mingyang Qian, and Yizhou Yu.
\newblock {Multi-source weak supervision for saliency detection}.
\newblock In \emph{CVPR}, 2019.

\bibitem[Zeng et~al.(2022)Zeng, Lin, and Patel]{zeng2022sketchedit}
Yu Zeng, Zhe Lin, and Vishal~M Patel.
\newblock {SketchEdit: Mask-Free Local Image Manipulation with Partial
  Sketches}.
\newblock In \emph{CVPR}, 2022.

\bibitem[Zhang et~al.(2024)Zhang, Jiang, Guo, Yan, Pan, Dong, Gao, and
  Li]{zhang2023personalize}
Renrui Zhang, Zhengkai Jiang, Ziyu Guo, Shilin Yan, Junting Pan, Hao Dong, Peng
  Gao, and Hongsheng Li.
\newblock {Personalize Segment Anything Model with One Shot}.
\newblock In \emph{ICLR}, 2024.

\bibitem[Zhang and Maire(2020)]{zhang2020self}
Xiao Zhang and Michael Maire.
\newblock {Self-Supervised Visual Representation Learning from Hierarchical
  Grouping}.
\newblock In \emph{NeurIPS}, 2020.

\bibitem[Zheng et~al.(2021)Zheng, Lu, Zhao, Zhu, Luo, Wang, Fu, Feng, Xiang,
  Torr, and Zhang]{zheng2021rethinking}
Sixiao Zheng, Jiachen Lu, Hengshuang Zhao, Xiatian Zhu, Zekun Luo, Yabiao Wang,
  Yanwei Fu, Jianfeng Feng, Tao Xiang, Philip H.~S. Torr, and Li Zhang.
\newblock {Rethinking Semantic Segmentation from a Sequence-to-Sequence
  Perspective with Transformers}.
\newblock In \emph{CVPR}, 2021.

\bibitem[Zhou et~al.(2016)Zhou, Khosla, Lapedriza, Oliva, and
  Torralba]{zhou2016learning}
Bolei Zhou, Aditya Khosla, Agata Lapedriza, Aude Oliva, and Antonio Torralba.
\newblock {Learning Deep Features for Discriminative Localization}.
\newblock In \emph{CVPR}, 2016.

\bibitem[Zhou et~al.(2022)Zhou, Yang, Loy, and Liu]{zhou2022learning}
Kaiyang Zhou, Jingkang Yang, Chen~Change Loy, and Ziwei Liu.
\newblock {Learning to Prompt for Vision-Language Models}.
\newblock \emph{IJCV}, 2022.

\bibitem[Zhou et~al.(2023)Zhou, Lei, Zhang, Liu, and Liu]{zhou2023zegclip}
Ziqin Zhou, Yinjie Lei, Bowen Zhang, Lingqiao Liu, and Yifan Liu.
\newblock {ZegCLIP: Towards Adapting CLIP for Zero-shot Semantic Segmentation}.
\newblock In \emph{CVPR}, 2023.

\end{thebibliography}
}

\end{document}